\documentclass[sigconf]{acmart}

\usepackage{booktabs} 
\usepackage{amsmath}
\usepackage{amsfonts}
\usepackage{makecell}
\usepackage{graphicx}
\usepackage{booktabs}
\usepackage{subcaption}
\usepackage[ruled]{algorithm2e}
\usepackage{multirow} 
\usepackage{amsfonts}
\usepackage{listings}
\usepackage{mathrsfs}
\usepackage{mathtools}
\usepackage{textcomp}
\usepackage{enumitem}
\newcolumntype{I}{!{\vrule width 1.5pt}}
\newlength\savedwidth

\newlength\savewidth
\newcommand\shline{\noalign{\global\savewidth\arrayrulewidth
                            \global\arrayrulewidth 1.5pt}%
                   \hline
                   \noalign{\global\arrayrulewidth\savewidth}}

\setcopyright{rightsretained}

\acmDOI{10.475/123_4}

\acmISBN{123-4567-24-567/08/06}

\acmYear{2019}
\copyrightyear{2019}

\acmArticle{4}
\acmPrice{15.00}

\begin{document}
\title{UrbanFM: Inferring Fine-Grained Urban Flows}

\author{Yuxuan Liang$^{1,2}$, Kun Ouyang$^{1}$, Lin Jing$^{2}$, Sijie Ruan$^{2,3}$, Ye Liu$^{1}$ \\ Junbo Zhang$^{3}$, David S. Rosenblum$^{1}$, Yu Zheng$^{2,3}$}
\affiliation{%
  \institution{$^{1}$School of Computing, National University of Singapore, Singapore
  \\ $^{2}$School of Computer Science and Technology, Xidian University, Xi'an, China 
  \\ $^{3}$JD Intelligent City Research \& JD Urban Computing Business Unit, Beijing, China}
}
\email{{yuxliang, amber_jing, sijieruan, msjunbozhang, msyuzheng}@outlook.com, {david,ouyangk,liuye}@comp.nus.edu.sg}

\renewcommand{\shortauthors}{Y. Liang et al.}

\begin{abstract}
Urban flow monitoring systems play important roles in smart city efforts around the world. However, the ubiquitous deployment of monitoring devices, such as CCTVs, induces a long-lasting and enormous cost for maintenance and operation. This suggests the need for a technology that can reduce the number of deployed devices, while preventing the degeneration of data accuracy and granularity. In this paper, we aim to infer the real-time and fine-grained crowd flows throughout a city based on coarse-grained observations. This task is challenging due to the two essential reasons: the spatial correlations between coarse- and fine-grained urban flows, and the complexities of external impacts. To tackle these issues, we develop a method entitled UrbanFM based on deep neural networks. Our model consists of two major parts: 1) an inference network to generate fine-grained flow distributions from coarse-grained inputs by using a feature extraction module and a novel distributional upsampling module; 2) a general fusion subnet to further boost the performance by considering the influences of different external factors. Extensive experiments on two real-world datasets, namely TaxiBJ and HappyValley, validate the effectiveness and efficiency of our method compared to seven baselines, demonstrating the state-of-the-art performance of our approach on the fine-grained urban flow inference problem.
\end{abstract}



\maketitle

\section{Introduction}
The fine-grained urban flow monitoring system is a crucial component of the information infrastructure system in smart cities, which lays the foundation for urban planning and various applications such as traffic management. To obtain data at a spatial fine-granularity, the system requires large amounts of sensing devices to be deployed in order to cover a citywide landscape. For example, thousands of piezoelectric sensors and loop detectors are deployed on road segments in a city to monitor vehicle traffic flow volumes in real time; many CCTVs are deployed ubiquitously for surveillance purposes and for obtaining real-time crowd flow data. With a large number of devices deployed, a high cost would be incurred due to the long-term operation (\textit{e.g.}, electricity and communication cost) and maintenance (\textit{e.g.}, on-site maintenance and warranty). A recent study showed that in Anyang, Korea, the annual operation and device maintenance fee for their smart city projects reached 100K USD and 400K USD respectively in 2015~\cite{idb-korea}. With the rapid development of smart cities on a worldwide scale, the cost of manpower and energy will become a prohibitive factor for the further intelligentization of the Earth. To reduce such expense, people require a novel technology which allows cutting the number of \textit{deployed} sensors while, most importantly, keeping the original data granularity unchanged. Therefore, how to approximate the original fine-grained information from available coarse-grained data (obtained from fewer sensors) becomes an urgent problem.
\begin{figure}[!h]
	\centering
	\includegraphics[width=0.47\textwidth]{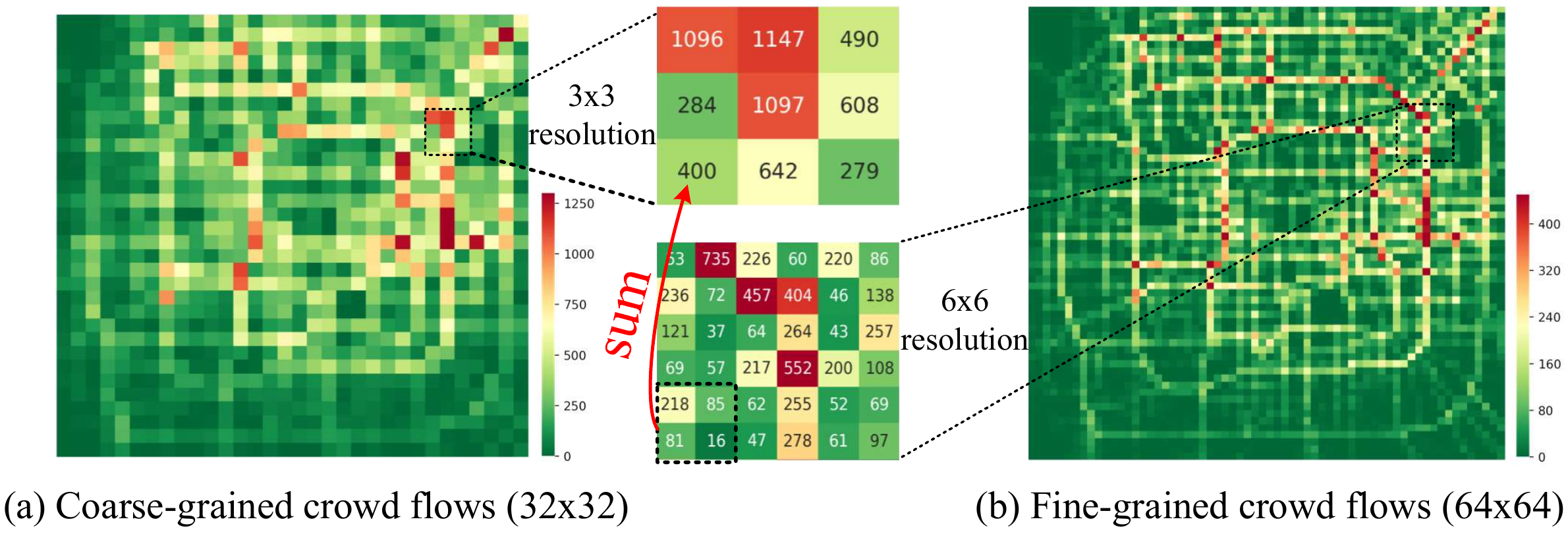}
	\vspace{-1em}
	\caption{\label{fig:intro} Traffic flows in two granularities in Beijing, where each grid denotes a region. Our aim is to infer (b) given the coarse-grained data source (a).}	
	\vspace{-1em}
\end{figure}

\begin{figure}[!t]
	\centering
	\includegraphics[width=0.49\textwidth]{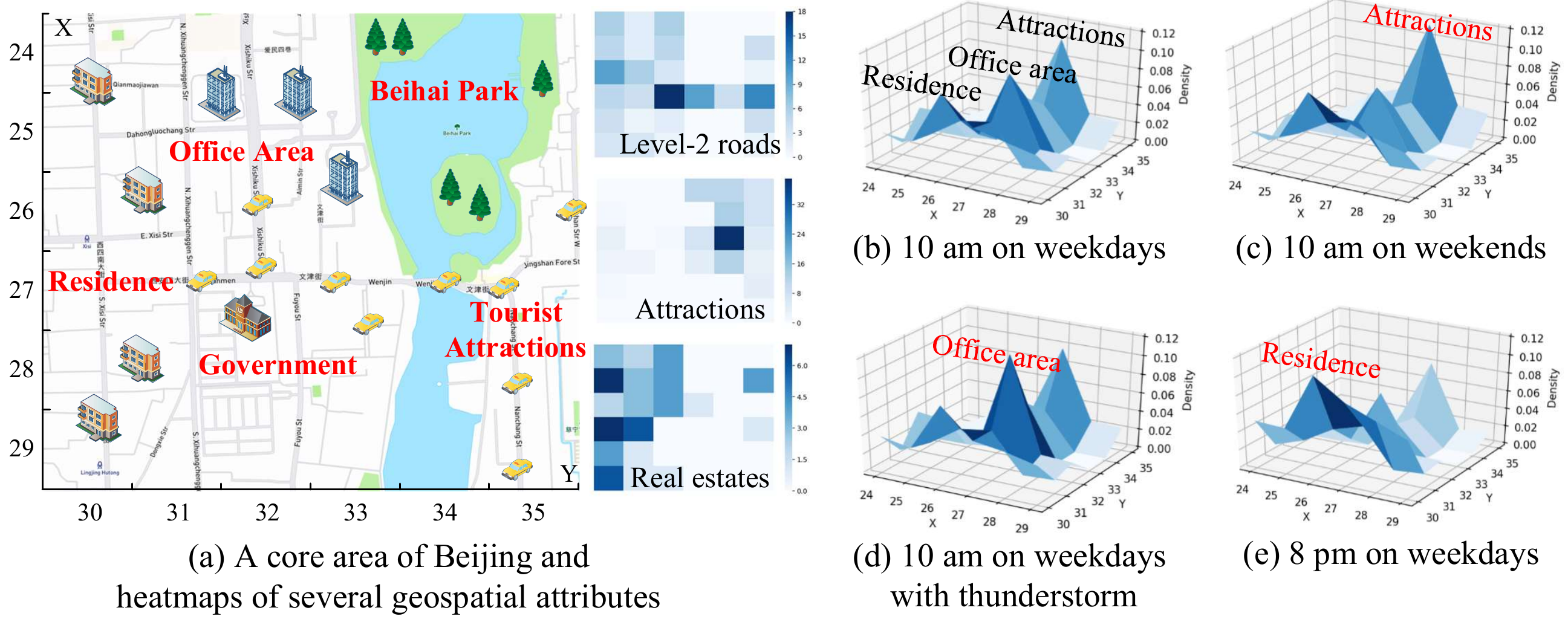}
	\vspace{-1.5em}
	\caption{\label{fig:ext_inf} The impact of external factors on the regional flow distributions. (a) We obtain Point of Interests (POIs) for different regions, and then categorize regions into different semantics according to the POI information. (b)-(d) depict the average flow distribution under various external conditions.}
	\vspace{-0.5em}
\end{figure}

Take monitoring traffic in a university campus as a regional example. We can reduce the number of the interior loop detectors and keep sensors only at the entrances of the campus to save cost. However, we still desire to recover the fine-grained flow distribution within the campus given only the coarse-grained information. In this paper, \textbf{our goal} is to infer the real-time and spatially fine-grained flow from observed coarse-grained data on a citywide scale with many other regions (as shown in Figure~\ref{fig:intro}(a)). This Fine-grained Urban Flow Inference (FUFI) problem, however, is very challenging due to the reasons as follows:

\begin{itemize}[leftmargin=*]
\item{\textbf{Spatial Correlations.} The fine-grained flow maps preserve spatial and structural correlations with the coarse-grained counterparts. Essentially, the flow volume in a coarse-grained superregion (\textit{e.g.}, the campus), is distributed to the constituent subregions (\textit{e.g.}, libraries) in the fine-grained situation. This implies a crucial structural constraint: the sum of the flow volumes in subregions strictly equals that of the corresponding superregion. Besides, the flow in a region can be affected by the flows in the nearby regions, which will impact the inference for the fine-grained flow distributions over subregions. Methods failing to capture these features would lead to a degenerated performance.}

\item{\textbf{External Factors.} The distribution of the flows in a given region is affected by various external factors, such as local weather, time of day and special events. To see the impact, we present a real-world study in an area of Beijing as shown in Figure~\ref{fig:ext_inf}(a). On weekdays, (b) shows more flows occur in the office area and attractions at 10 a.m. as compared to at 8 p.m. where residences witness much higher flow density than the others (see (e)); on weekends, however, (c) demonstrates that people tend to present in a park area in the morning. It corresponds to our common sense that people go out for work in the morning, to attractions for relaxation in the weekend and return home at night. Besides, (d) shows that people keen to move to indoor areas instead of the outdoor park under storms. These observations evince that regions with different semantics present different flow distributions in respect of different external factors. Moreover, these external factors can intertwine and thus influence the actual distribution in a very complicated manner.}

\end{itemize}

The FUFI problem can be cast as a mapping problem which maps data of low information entropy to that of high information entropy. Sharing the same nature with FUFI, recent studies~\cite{dong2016srcnn,lim2017enhanced,ledig2017srgan} in image super-resolution (SR) have presented techniques for recovering high-resolution images from low-resolution images, which has motivated applications in other fields, such as meteorology~\cite{vandal2017deepsd}. Nevertheless, due to the aforementioned challenges, the simple application of these techniques to FUFI is infeasible and thus requires a careful redesign of the model architecture.

To this end, we present \textbf{Urban} \textbf{F}low \textbf{M}agnifier (\textbf{UrbanFM}), a deep neural network model which learns to infer fine-grained urban flows under the supervised-learning paradigm. Following related techniques~\cite{dong2016srcnn,lim2017enhanced,ledig2017srgan,vandal2017deepsd}, we assume a number of fine-grained data are available to bootstrap our solution\footnote{The original data can be obtained through previously deployed sensors or from crowd-sourcing.}. The key contributions of this paper lie in the following aspects:
\begin{itemize}[leftmargin=*]
\item{We present the first attempt to formalize the fine-grained urban flow inference problem with identification of the problem specificities and relevant challenges.}
\item{We design an inference network to handle the spatial correlations. The inference network employs a convolutional network-based feature extraction module to address the nearby region influence. More importantly, it leverages a   \textit{distributional upsampling} module with a novel and parameter-free layer entitled $N^2$-\textit{Normalization} to impose the structural constraint on the model, by converting the learning focus from directly generating flow volumes (as in related arts) to inferring the actual flow distribution.}
\item{We design an external factor fusion subnet to account for all complex external influences at once. The subnet generates an integrated, high-level representation for external factors. The hidden features are then fed into different levels of the inference network (\emph{i.e.}, coarse- and fine-grained levels) to enhance the inference performance.}
\item{We process, analyze, and experiment in two real-world urban scenarios, including the taxi flows with a metropolitan coverage and the human flows within a touristic district respectively. Our experimental results verify the significant advantages of UrbanFM over five state-of-the-art and two heuristical methods in both effectiveness and efficiency. Moreover, the experiments from multiple prospectives validate the rationale for different components of the model. We have released the code, sample data and demo for public use\footnote{https://github.com/yoshall/UrbanFM}.}
\end{itemize}

\section{Formulation}\label{sec:form}
In this section, we first define the notations and then formulate the problem of Fine-grained Urban Flow Inference (FUFI).

\noindent\textbf{Definition 1 (Region)} As shown in Figure~\ref{fig:intro}, we partition an area of interest (\emph{e.g.}, a city) evenly into a $I\times J$ grid map based on the longitude and latitude, where a grid denotes a region~\cite{zhang2017deep}. Partitioning the city into smaller regions (\emph{i.e.}, using larger $I,J$) suggests that we can obtain flow data with more details, which results in a more fine-grained flow map.

\noindent\textbf{Definition 2 (Flow Map)} Let $\mathbf{X}\in\mathbb{R}^{I\times J}_{+}$ represent a flow map of a particular time, where each entry $x_{i,j}\in \mathbb{R}_{+}$ denotes the flow volume of the instances (\emph{e.g.}, vehicle, people, etc.) in region $(i,j)$.

\noindent\textbf{Definition 3 (Superregion \& Subregion)} In our FUFI problem, a coarse-grained grid map indicates the data granularity  we can observe upon sensor reduction. It is obtained by integrating nearby grids within an $N$-by-$N$ range in a fine-grained grid map given a scaling factor N. Figure~\ref{fig:intro} illustrates an example when $N=2$. Each coarse-grained grid in Figure~\ref{fig:intro}(a) is composed of $2\times 2$ smaller grids from Figure~\ref{fig:intro}(b). We define the aggregated larger grid as \textit{superregion}, and its constituent smaller regions as \textit{subregions}. Note that with this setting, the superregions do not share subregions. Hence, the structure between superregions and the corresponding subregions indicates a special \textit{structural constraint} in FUFI.

\begin{figure*}[!t]
	\centering
	\includegraphics[width=0.98\textwidth]{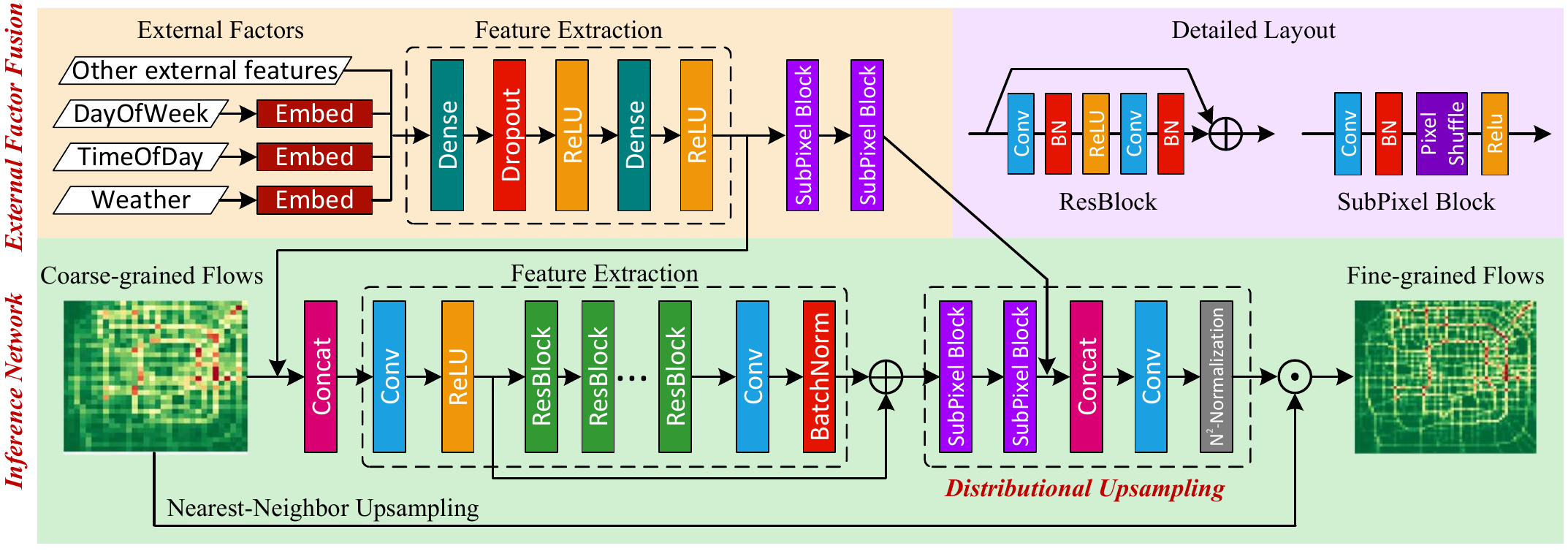}
	\vspace{-1em}
	\caption{\label{fig:framework}The UrbanFM framework for $4\times$ upscaling ($N=4$). $\oplus$ denotes addition and $\odot$ denotes Hadamard product. Note that our framework allows an arbitrary integer upscaling factor, not limited to the power of 2.}
	\vspace{-1em}
\end{figure*}

\noindent\textbf{Definition 4 (Structural Constraint)} The flow volume $x^c_{i,j}$ in a superregion of the coarse-grained grid map and the flows $x^f_{i',j'}$ in the corresponding subregions of the fine-grained counterpart obey the following equation:
\begin{equation}
	\label{eqn:constrain}
	x^c_{i,j} = \sum_{i',j'}{x^f_{i',j'}} \quad s.t. \lfloor{\frac{i'}{N}}\rfloor=i,\lfloor{\frac{j'}{N}}\rfloor=j.
\end{equation}
For simplicity, $i=1,2,\dots,I$ and $j=1,2,\dots,J$ in our paper unless otherwise specified.

\noindent\textbf{Problem Statement (Fine-grained Urban Flow Inference)}

\noindent Given a upscaling factor $N\in Z$ and a coarse-grained flow map $\mathbf{X}^c\in\mathbb{R}^{I\times J}_{+}$, infer the fine-grained counterpart $\mathbf{X}^f\in\mathbb{R}^{NI\times NJ}_{+}$.

\section{Methodology}

Figure~\ref{fig:framework} depicts the framework of UrbanFM, which consists of two main components for conducting the structurally constrained inference and capturing complex external influence, respectively. The inference network takes the coarse-grained flow map $\mathbf{X}^c$ as input, then extracts high-level features across the whole city by leveraging deep residual networks~\cite{he2016deep}. Taking extracted features as a priori knowledge, the \textit{distributional upsampling} module outputs a flow distribution over subregions with respect to each superregion by introducing a dedicated $N^2$-\textit{Normalization} layer. Finally, the Hadamard product of the inferred distribution with the upsampled coarse-grained flow map gives the fine-grained flow map $\tilde{\mathbf{X}}_f$ as the network output. In the external factor fusion branch, we leverage embeddings and a dense network to extract pixel-wise external features in both coarse and fine granularity. The integration of external and flow features enables UrbanFM to exhibit fine-grained flow inference more effectively. In this section, we articulate the key designs for the two components, as well as the optimization scheme on network training.

\vspace{-1mm}
\subsection{Inference Network\label{sec:infnet}}
Inference network aims to produce the fine-grained flow distribution over subregions from a coarse-grained input. We follow the general procedure in SR methods, which is composed of two phases: 1) feature extraction; 2) inference upon upsampling.

\vspace{-1mm}
\subsubsection{Feature Extraction}
In the input stage, we use a convolutional layer (with $9\times 9$ filter size, $F$ filters) to extract low-level features from the given coarse-grained flow map $\mathbf{X}^c$, and perform the first stage fusion if external features are provided. Then $M$ Residual Blocks with identical layout take the (fused) low-level feature maps as input and then construct high-level feature maps. The residual block layout, as shown on the top right of Figure~\ref{fig:framework}, follows the guideline in \cite{ledig2017srgan}, which contains two convolutional layers ($3 \times 3$, $F$) followed by Batch Normalization~\cite{ioffe2015batch}, with an intermediate ReLU~\cite{hahnloser2000digital} function to introduce non-linearity.

Since we utilize a fully convolutional architecture, the reception field grows larger as we stack the network deeper. In other words, each pixel at the high-level feature map will be able to capture distant or even citywide dependencies. Moreover, we use another convolutional layer ($3\times 3$, $F$) followed by batch normalization to guarantee feature extraction. Finally, drawing from the intuition that the output flow distribution would exhibit region-to-region dependencies to the original $\mathbf{X}^c$, we employ a skip connection to introduce identity mapping~\cite{he2016identity} between the low-level features and high-level features, building an information highway skipping over the residual blocks to allow efficient gradient back-propagation.

\vspace{-1mm}
\subsubsection{Distributional Upsampling}
In the second phase, the extracted features first go through $n$ sub-pixel blocks to perform an $N=2^n$ upscaling operation which produces a hidden feature $\mathbf{H}^f \in \mathbb{R}^{F\times~NI\times~NJ}$. The sub-pixel block, as illustrated in Figure~\ref{fig:framework}, leverages a convolutional layer ($3\times 3$, $F\times2^2$) followed by batch normalization to extract features. Then it uses a PixelShuffle layer~\cite{shi2016espcn} to rearrange and upsample the feature maps to $2\times$ size and applies a ReLU activation at the end. After each sub-pixel block, the output feature maps grow 2 times larger with the number of channels unchanged. A convolutional layer ($9\times 9$, $F_o$) is applied post-upsampling, which maps $\mathbf{H}^f$ to a tensor $\mathbf{H}^f_o \in \mathbb{R}^{F_o\times~NI\times~NJ}$. Note that $F_o=1$ in our case. In SR tasks, $\mathbf{H}^f_o$ is usually the final output for the recovered image with super-resolution. However, the structural constraint essential to FUFI has not been considered.
%
%

In order to impose the structural constraint on the network, one straightforward manner is to add a \emph{structural loss} $L_s$ as a regularization term to the loss function:
%
\begin{equation}
	L_s = \sum_{i,j}{\bigg\lVert x^c_{i,j} - \sum_{i',j'}{\tilde{x}^f_{i',j'}}\bigg\rVert_F} \quad s.t. \lfloor{\frac{i'}{N}}\rfloor=i,\lfloor{\frac{j'}{N}}\rfloor=j.
\end{equation}

However, simply applying $L_s$ does not improve the model performance, as we will demonstrate in the experiment section. Instead, we design a $N^2$-\textit{Normalization} layer, which outputs a \textit{distribution} over every patch of $N$-by-$N$ subregions with regard to the respective superregion. To achieve this, we reformulate Equation~\ref{eqn:constrain} as in the following:
%
\begin{equation}
	\begin{aligned}
	x^c_{i,j} = \sum_{i',j'}{\alpha^{\;}_{i',j'}x^c_{i,j}}\\
	s.t. \sum{\alpha_{i',j'}}=1, \;\alpha\in~\mathbb{R}_{+}, \lfloor{\frac{i'}{N}}\rfloor&=i,\lfloor{\frac{j'}{N}}\rfloor=j.
	\end{aligned}
\end{equation}
The flow volume in each subregion is now expressed as a \textit{fraction} of that in the superregion, \emph{i.e.}, $x^f_{i',j'}=\alpha^{\;}_{i'j'}x^c_{i,j}$, and we can treat the fraction as a probability. This allows us to interpret the network output in a meaningful way: the value in each pixel states how likely the overall flow will be allocated to the subregion $(i',j')$. By reformulation, we shift our focus from directly generating the fine-grained flow to generating the flow distribution. This essentially changes the network learning target and thus diverges from the traditional SR literature. To this end, we present the $N^2$-\textit{Normalization} layer: $N^2$-\textit{Normalization}$(\mathbf{H_o^f}) = \mathbf{H}^f_\pi$, such that
%
\begin{equation}
	\mathbf{H}^{f}_{\pi,(i,j)}=\mathbf{H}^{f}_{o,(i,j)}/\sum_{i'=(\lfloor{i/N}\rfloor-1)*N+1, \atop j'=(\lfloor{j/N}\rfloor-1)*N+1}^{i'=\lfloor{i/N}\rfloor*N, \atop j'=\lfloor{j/N}\rfloor*N}{\mathbf{H}^{f}_{o,(i',j')}}
	\label{eqn:n2}
\end{equation}

$N^2$-Normalization layer induces no extra parameters for the network. Moreover, it can be easily implemented within a few lines of code (see Algorithm 1). Also, the operations can be fully paralleled and automatically differentiated in runtime. Remarkably, this reformulation release the network from concerning varying output scales and enable it to focus on producing a probability within $[0,1]$ constraint.
\begin{algorithm}[!t]
\SetAlgoLined
 \KwIn{x, scale\_factor, $\epsilon$}
 \KwOut{out} 
 \tcp{x: an input feature map} \vspace{-0.5mm}
 \tcp{scale\_factor: the upscaling factor} \vspace{-0.5mm}
 \tcp{$\epsilon$: a small number for numerical stability} \vspace{-0.5mm}
 \tcp{out: the structural distributions}
 \vspace{2mm}
 sum = SumPooling(x, scale\_factor)\;
 sum = NearestNeighborUpsampling(sum, scale\_factor)\;
 out \hspace{1mm}= x $\oslash$ (sum+$\epsilon$) // \texttt{element wise division} 
 \caption{$N^2$-Normalization}
 \label{alg:n2}
\end{algorithm}

Finally, we upscale $\mathbf{X}^c$ using nearest-neighbor upsampling\footnote{https://en.wikipedia.org/wiki/Nearest-neighbor\_interpolation} with the scaling factor $N$ to obtain $\mathbf{X}^c_{up}\in\mathbb{R}^{NI\times NJ}_{+}$ and then generate the fine-grained inference by $\tilde{\mathbf{X}}^f=\mathbf{X}^c_{up}\odot\mathbf{H^f_\pi}$.

\subsection{External Factor Fusion}
External factors, such as weather, can have a complicated and vital influence on the flow distribution over the subregions. For instance, even if the total population in town remains stable over time, under storming weather people tend to move from outdoor regions to indoor regions. When different external factors entangle, the actual impact on the flow becomes implicit however unneglectable. Thereby, we design a subnet to handle those impacts \textit{all at once}.

Particularly, we first separate the available external factors into two groups, \emph{i.e.}, continuous and categorical features. Continuous features including temperature and wind speed are directly concatenated to be a vector $\mathbf{e}_{con}$. As shown in Figure \ref{fig:framework}, categorical features include the day of week, the time of the day and weather (\emph{e.g}, sunny, rainy). Inspired by previous studies \cite{liang2018geoman}, we transform each categorical attribute into a low-dimensional vector by feeding them into different embedding layers separately, and then concatenate those embeddings to construct the categorical vector $\mathbf{e}_{cat}$. Then, the concatenation of $\mathbf{e}_{con}$ and $\mathbf{e}_{cat}$ gives the final embeddings for external factors, \emph{i.e.}, $\mathbf{e}=[\mathbf{e}_{con};\mathbf{e}_{cat}]$.

Once we get the concatenation vector $\mathbf{e}$, we feed it into a feature extraction module, whose structure is depicted in Figure~\ref{fig:framework}. By using dense layers, different external impacts are compounded to construct a hidden representation, which models the complicated interaction. The module provides two outputs: the coarse-grained feature maps $\mathbf{H}^c_e$ and the fine-grained feature maps $\mathbf{H}^f_e$, where $\mathbf{H}^f_e$ is obtained by passing $\mathbf{X}^c_e$ through $n$ sub-pixel blocks which are similar to the ones in the inference network. Intuitively, $\mathbf{H}^c_e$ ($\mathbf{H}^f_e$) is the spatial encoding for $\mathbf{e}$ in coarse-grained (fine-grained) setting, modeling how each superregion (subregion) individually responds to the external changes. Therefore we concatenate $\mathbf{H}^c_e$ with $\mathbf{X}^c$, and $\mathbf{H}^f_e$ with $\mathbf{H}^f$ to the inference network. The early fusion of $\mathbf{H}^c_e$ and $\mathbf{X}^c$ allows the network to learn to extract a high-level feature describing not only the citywide flow, but also the external influences. Besides, the fine-grained $\mathbf{H}^f_e$ carries the external information all the way to the rear of the inference network, playing a similar role as an information highway, which prevents information perishing in the deep network.

\subsection{Optimization}
UrbanFM provides an end-to-end mapping from coarse-grained input to fine-grained output, which is differentiable everywhere. Therefore, we can learned the network through auto back propagation, by providing training pairs ($\mathbf{X}^c, \mathbf{X}^f$) and calculating empirical loss between ($\mathbf{X}^f,\tilde{\mathbf{X}}^f$), where $\mathbf{X}^f$ is the ground truth and $\tilde{\mathbf{X}}^f$ is the outcome inferred by our network. Pixel-wise Mean Square Error (MSE) is a widely used cost function in many tasks, and we employ the same in this work as follows:
\begin{equation}
L(\Omega)=\lVert{\mathbf{X}^f-\tilde{\mathbf{X}}^f}\rVert_F^2
\end{equation}
where $\Omega$ denotes the set of parameters in UrbanFM. Noted that $M$ and $F$ are the two main hyperparameters controlling the learning ability as well as the parameter size of the network. We experiment with different hyperparameter settings in the next section.


\section{Experiments}\label{sec:exp}
The focus of our experiments is on examining the capacity of our model in a citywide scenario. Therefore, we conduct extensive experiments using taxi flows in Beijing to comprehensively test the model from different aspects. In addition, we conduct further experiments in a theme park, namely Happy Valley, to show the model adaptivity on a relatively small area. 

\subsection{Experimental Settings}
\subsubsection{Datasets}
\begin{figure}[h!]
	\centering
	\includegraphics[width=0.45\textwidth]{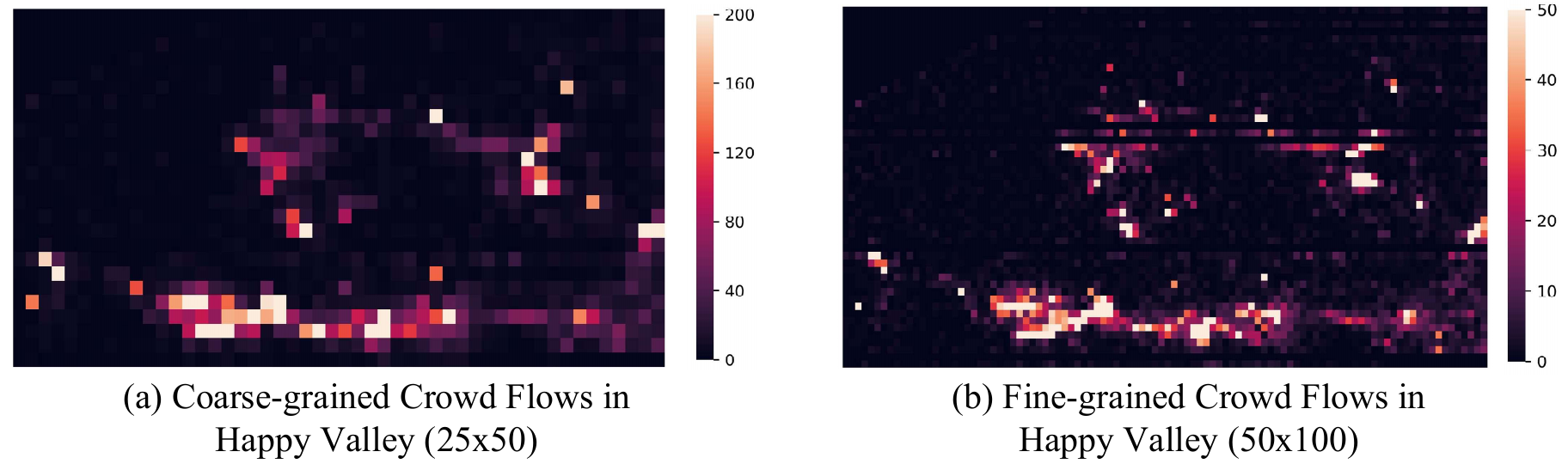}
	\caption{Visualization of crowd flows in HappyValley.}
	\vspace{-1em}	
	\label{fig:dataset}
\end{figure}

Table \ref{tab:dataset} details the two datasets we use, namely TaxiBj and HappyValley, where each dataset contains two sub-datasets: urban flows and external factors. Since a number of fine-grained flow data are available as ground truth, in this paper, we obtain the coarse-grained flows by aggregating subregion flows from the fine-grained counterparts.
\begin{itemize}[leftmargin=*]
	\item \textbf{TaxiBJ}: This dataset, which is published by \citeauthor{zhang2017deep} \cite{zhang2017deep}, indicates the taxi flows traveling throughout Beijing. As depicted in Figure \ref{fig:intro}, the studied area is split into 32$\times$32 grids, where each grid reports the coarse-grained flow information every 30 minutes within four different periods: P1 to P4 (detailed in Table \ref{tab:dataset}). Here, we utilize the coarse-grained taxi flows to infer fine-grained flows with 4$\times$ resolution ($N=4$). In our experiment, we partition the data into non-overlapping training, validation and test data by a ratio of 2:1:1 respectively for each period. For example, in P1 (7/1/2013-10/31/2013), we use the first two-month data as the training set, the next month as the validation set, and the last month as the test set.
	\item \textbf{HappyValley}: We obtain this dataset by crawling from an open website\footnote{heat.qq.com} which provides hourly gridded crowd flow observations for a theme park named Beijing Happy Valley, with a total 1.25$\times10^5 m^2$ area coverage, from 1/1/2018 to 10/31/2018. As shown in Figure~\ref{fig:dataset}, we partition this area with 25$\times$50 uniform grids in coarse-grained setting, and target a fine granularity at 50$\times$100 with an upscaling factor $N=2$. Note that in this dataset, one special external factor is the ticket price, including day prices and night prices, which are obtained from the official account in WeChat. Regarding the smaller area, crowd flows exhibits large variance across samples given the 1-hour sampling rate. Thus, we use a ratio of 8:1:1 to split training, validation and test set to provide more training data.
\end{itemize}
\begin{table}[!h]
	\centering
	\caption{Dataset Description.}
	\vspace{-0.8em}
	\tabcolsep=1.2mm
	  \begin{tabular}{lll}
	  \shline
	  \textbf{Dataset} & \textbf{TaxiBJ} & \textbf{HappyValley} \\
	  \hline
	  \multirow{4}[1]{*}{Time span} & P1: 7/1/2013-10/31/2013 &  \\
			& P2: 2/1/2014-6/30/2014 & 1/1/2018- \\
			& P3: 3/1/2015-6/30/2015 & 10/31/2018 \\
			& P4: 11/1/2015-3/31/2016 &  \\
	  \midrule
	  Time interval & 30 minutes & 1 hour \\
	  Coarse-grained size & 32$\times$32 & 25$\times$50 \\
	  Fine-grained size & 128$\times$128 & 50$\times$100 \\
	  Upscaling factor ($N$) & 4     & 2 \\
	  \midrule
	  \multicolumn{3}{l}{\textbf{External factors (meteorology, time and event)}} \\
	  Weather (\emph{e.g.}, Sunny) & 16 types & 8 types \\
	  Temperature/\textcelsius & [-24.6, 41.0] & [-15.0, 39.0] \\
	  Wind speed/mph & [0, 48.6] & [0.1, 15.5] \\
	  \# Holidays & 41    & 33 \\
	  Ticket prize/\textyen  & /     & [29.9, 260] \\
	  \shline
	  \end{tabular}%
	\label{tab:dataset}%
  \end{table}%

\subsubsection{Evaluation Metrics}

We use three common metrics for urban flow data to evaluate the model performance from different facets. Specifically, Root Mean Square Error (RMSE) is defined as:
\begin{equation*}
\small
	RMSE = \sqrt{\frac{1}{z}\sum_{s=1}^{z}{\bigg\lVert \mathbf{X}^f_s-\tilde{\mathbf{X}}^f_s} \bigg\rVert^2_F},
\end{equation*} 
where $z$ is the total number of samples, $\tilde{\mathbf{X}}^f_s$ is $s$-th the inferred value and $\mathbf{X}^f_s$ is corresponding ground truth. Mean Absolute Error (MAE) and Mean Absolute Percentage Error (MAPE) are defined as: $MAE = \frac{1}{z}\sum_{s=1}^{z}{\lVert{\mathbf{X}^f_s-\tilde{\mathbf{X}}^f_s}}\rVert_{1,1}$ and $MAPE = \frac{1}{z}\sum_{s=1}^{z}{\lVert{(\mathbf{X}^f_s-\tilde{\mathbf{X}}^f_s}) \oslash \mathbf{X}^f_s} \rVert_{1,1}$. In general, RMSE favors spiky distributions, while MAE and MAPE focus more on the smoothness of the outcome. Smaller metric scores indicate better model performance.

\begin{table*}[!t]
  \centering
  \caption{\label{tab:main_bj}Results comparisons on TaxiBJ over different time spans (P1-P4).}
  \vspace{-0.8em}
  \tabcolsep=0.22cm
    \begin{tabular}{c|ccc|ccc|ccc|ccc}
    \shline
    \multirow{2}*{Methods} & \multicolumn{3}{c|}{P1} & \multicolumn{3}{c|}{P2} & \multicolumn{3}{c|}{P3} & \multicolumn{3}{c}{P4} \\
\cline{2-13}           & RMSE  & MAE   & MAPE  & RMSE  & MAE   & MAPE  & RMSE  & MAE   & MAPE  & RMSE  & MAE   & MAPE \\
    \hline
    MEAN  & 20.918 & 12.019 & 4.469 & 20.918 & 12.019 & 5.364 & 27.442 & 16.029 & 5.612 & 19.049 & 11.070 & 4.192 \\

    HA    & 4.741 & 2.251 & 0.336 & 5.381 & 2.551 & 0.334 & 5.594 & 2.674 & 0.328 & 4.125 & 2.023 & 0.323 \\
    \hline
    SRCNN & 4.297 & 2.491 & 0.714 & 4.612 & 2.681 & 0.689 & 4.815 & 2.829 & 0.727 & 3.838 & 2.289 & 0.665 \\

    ESPCN & 4.206 & 2.497 & 0.732 & 4.569 & 2.727 & 0.732 & 4.744 & 2.862 & 0.773 & 3.728 & 2.228 & 0.711 \\

    DeepSD & 4.156 & 2.368 & 0.614 & 4.554 & 2.612 & 0.621 & 4.692 & 2.739 & 0.682 & 3.877 & 2.297 & 0.652 \\
    \hline
    VDSR  & 4.159 & 2.213 & 0.467 & 4.586 & 2.498 & 0.486 & 4.730 & 2.548 & 0.461 & 3.654 & 1.978 & 0.411 \\

    SRResNet & 4.164 & 2.457 & 0.713 & 4.524 & 2.660 & 0.688 & 4.690 & 2.775 & 0.717 & 3.667 & 2.189 & 0.637 \\

    \hline
    UrbanFM-ne & 4.015 & 2.047 & 0.332 & 4.386 & 2.258 & 0.320 & 4.559 & 2.352 & 0.316 & 3.559 & 1.845 & 0.309 \\

    \textbf{UrbanFM} & \textbf{3.950} & \textbf{2.011} & \textbf{0.327} & \textbf{4.329} & \textbf{2.224} & \textbf{0.313} & \textbf{4.496} & \textbf{2.318} & \textbf{0.315} & \textbf{3.501} & \textbf{1.815} & \textbf{0.308} \\
    \shline

    \end{tabular}%
\end{table*}%

\subsubsection{Baselines}
We compare our proposed model with seven baselines that belong to the following three classes: (1) Heuristics. (2) Image super-resolution. (3) Meteorological super-resolution. The first two methods are designed by us based on intuition or empirical knowledge, while the next four methods are previously and currently state-of-the-art methods for single image super-resolution. The last method is the state of the art on statistical downscaling for climate data. We detail them as follows:

\begin{itemize}[leftmargin=*]
  \item \textbf{Mean Partition (Mean)}: We evenly distribute the flow volume from each superregion in a coarse-grained flow map to the $N^2$ subregions, where $N$ is the upscaling factor.
  \item \textbf{Historical Average (HA)}: Similar to distributional upsampling, HA treats the value over each subregion a fraction of the value in the respective super region, where the faction is computed by averaging all training data.
  \item \textbf{SRCNN} \cite{dong2016srcnn}: SRCNN presents the first successful introduction of convolutional neural networks (CNNs) into the SR problems. It consists of three layers: patch extraction, non-linear mapping and reconstruction. Filters of spatial sizes $9 \times 9$, $5 \times 5$, and $5 \times 5$ were used respectively. The number of filters in the two convolutional layers are 64 and 32 respectively. In SRCNN, the low-resolution input is upscaled to the high-resolution space using a single filter (commonly bicubic interpolation) before reconstruction.
  \item \textbf{ESPCN} \cite{shi2016espcn}: Bicubic interpolation used in SRCNN is a special case of the deconvolutional layer. To overcome the low efficiency of such deconvolutional layer, Efficient Sub-Pixel Convolutional Neural Network (ESPCN) employs a sub-pixel convolutional layer that aggregates the feature maps from LR space and builds the SR image in a single step.
  \item \textbf{VDSR} \cite{kim2016vdsr}: Since both SRCNN and ESPCN follow a three-stage architecture, they have several drawbacks such as slow convergence speed and limited representation ability. Inspired by the VGG-net, \citeauthor{kim2016vdsr} presents a Super-Resolution method using Very Deep neural networks with depth up to 20. This study suggests that large depth is necessary for the task of SR.
  \item \textbf{SRResNet} \cite{ledig2017srgan}: SRResNet enhances VDSR by using the residual architecture presented by~\citeauthor{he2016deep}\cite{he2016deep}. The residual architecture allows one to stack a much larger number of network layers, which bases many benchmark methods in computer vision tasks.
  \item \textbf{DeepSD} \cite{vandal2017deepsd}: DeepSD is the state-of-the-art method on statistical downscaling\footnote{Downscaling means obtaining higher resolution image in meteorology~\cite{vandal2017deepsd} while the opposite in computer graphics~\cite{dong2016srcnn}.} for meteorological data. It basically exploits SRCNN for downscaling for an intermediate level, and performs further downscaling by simply stacking more SRCNNs. This method, however, would inherently require much more parameters compared with our method.
\end{itemize}

\subsubsection{Variants}
To evaluate each component of our method, we also compare it with different variants of UrbanFM:
\begin{itemize}[leftmargin=*]
	\item \textbf{UrbanFM-ne}: We simply remove the external factor fusion subnet from our method, which can help reveal the significance of this component.
	\item \textbf{UrbanFM-sl}: Upon removing the external subnet, we further replace distributional upsampling module by using sub-pixel blocks and $L_s$ to consider the structural constraint in this variant. 
	\end{itemize}

\subsubsection{Training Details \& Hyperparameters}

Our model, as well as the baselines, are completely implemented by PyTorch  with one TITAN V GPU. We leverage Adam~\cite{kingma2014adam}, an algorithm for stochastic gradient descent, to perform network training with learning rate $lr=1e-4$ and batch size being 16. We also apply a staircase-like schedule by halving the learning rate every 20 epochs, which allows smoother search near the convergence point. In the external subnet, there are 128 hidden units in the first dense layer with dropout rate 0.3, and $I \times J$ hidden units in the second dense layer. We embed DayOfWeek to $\mathbb{R}^2$, HourOfDay to $\mathbb{R}^3$ and weather condition to $\mathbb{R}^3$. Besides, for VDSR and SRResNet, we use the default settings in their paper. Since SRCNN, ESPCN and DeepSD perform poorly with default settings, we test different hyperparameters for them and finally use 768 and 384 as the number of filters in their two convolutional layers respectively. See more details in Appendix.

\subsection{Results on TaxiBJ}
\noindent\textbf{Model Comparison}

\noindent In this subsection, we compare the model effectiveness against the baselines. We report the result of UrbanFM with $M$-$F$ being 16-128 as our default setting. Further experiments regarding different $M$-$F$ will be discussed later. Likewise, we postpone the result of UrbanFM-sl to the next experiment for a more detailed study.

Table~\ref{tab:main_bj} summarizes the experimental results on TaxiBJ. We have the following observations: (1) The UrbanFM and its variant outperform all baseline methods in all three metrics over all time spans (P1-P4). Take SRRestNet for example. UrbanFM advances it by 4.5\%, 17.0\% and 54.1\% for RMSE, MAE and MAPE on average, where UrbanFM-ne also advances by 3.0\%, 15.6\% and 53.6\% respectively. The advance of UrbanFM-ne over all baselines indicates that the distribution upsampling in our inference network plays a leading role in improving the inference performance; the advance of UrbanFM over UrbanFM-ne supports that the combination with external subnet indeed enhances the model by incorporating external factors. (2) Image super-resolution methods outdo the heuristic method HA on RMSE while show deteriorate scores on MAE and MAPE. This can be attributed to two reasons: first, neural network methods are dedicated to performing well on RMSE as it is the training objective; second, HA preserves the spatial correlation for fine-grained flow maps while the others fail to do so. This again emphasizes the importance of preserving the structural constraint. A piece of further evidence can be seen from the comparison between UrbanFM-ne and SRResNet, where the former model has a similar structure as SRResNet except the distributional upsampling module, which makes it surpass its counterpart. Due to the similarity of model architecture, we select SRResNet as the baseline model for subsequent studies over different UrbanFM components.

\begin{figure}[!b]
	\centering
	\vspace{-1em}
	\begin{subfigure}[b]{0.23\textwidth}
		\includegraphics[width=\textwidth]{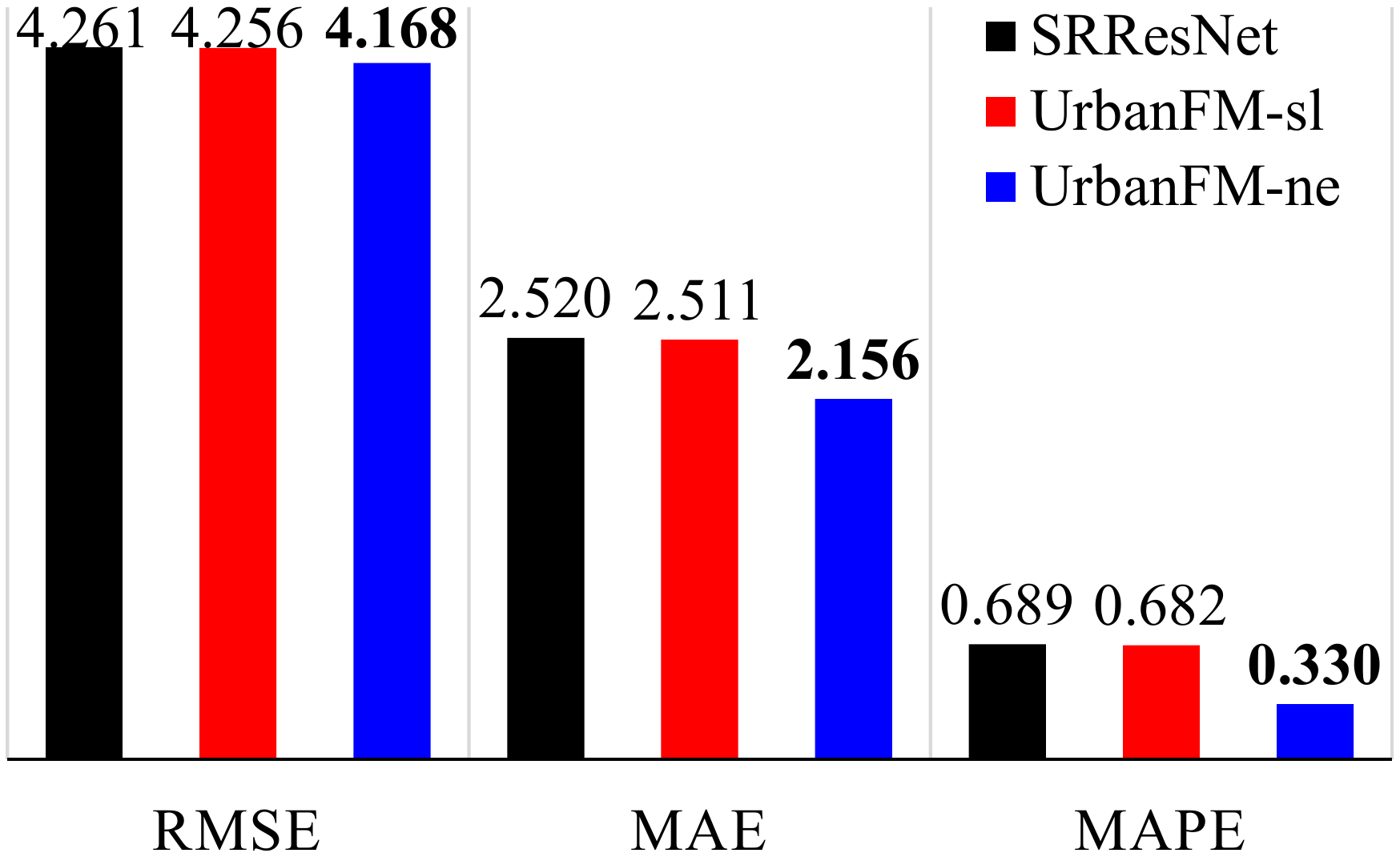}
		\caption{16-64 setting}
		\label{fig:hiloss1}
	\end{subfigure}
	\begin{subfigure}[b]{0.23\textwidth}
		\includegraphics[width=\textwidth]{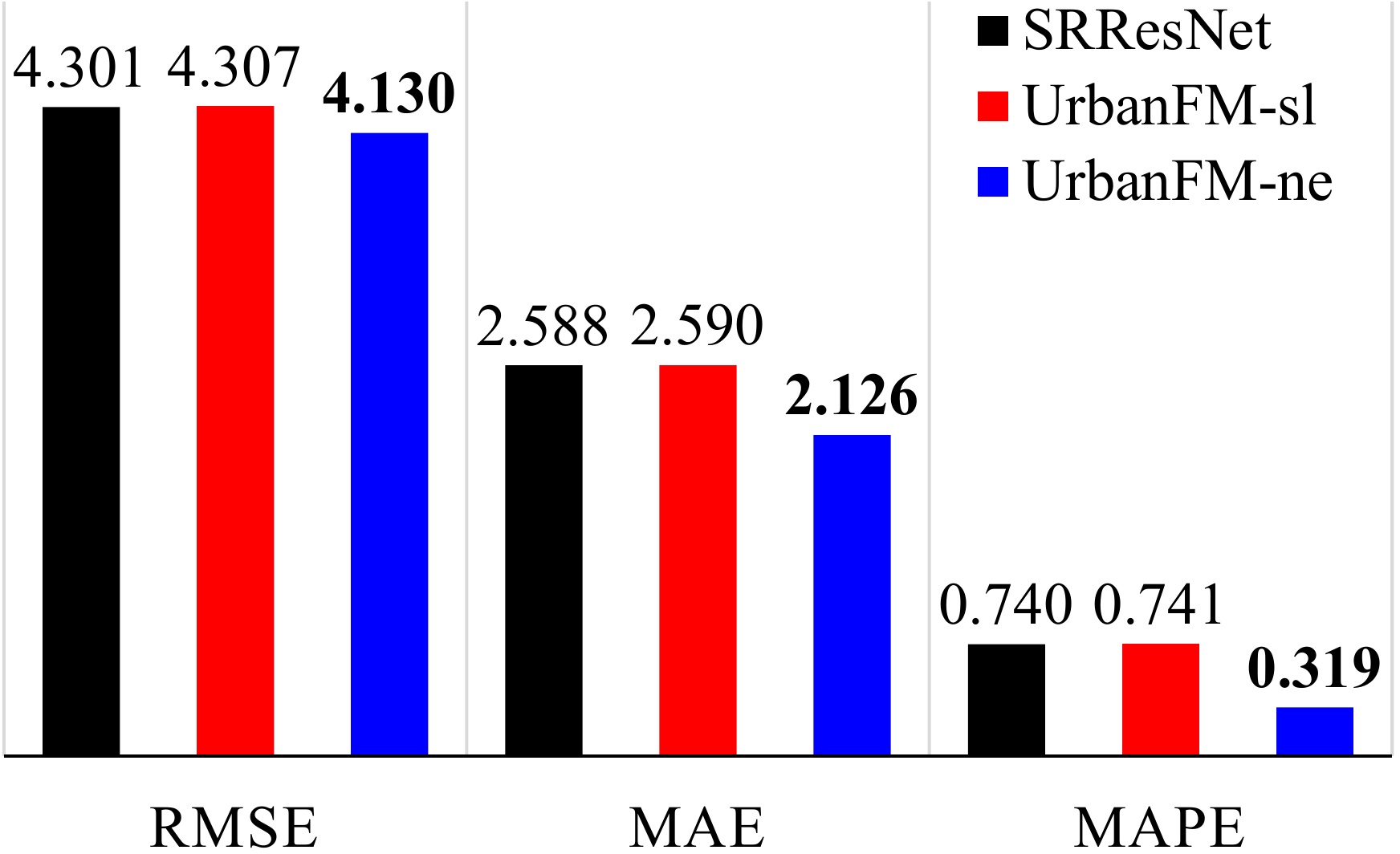}
		\caption{16-128 setting}
		\label{fig:hiloss2}
	\end{subfigure}
	\vspace{-1em}
	\caption{Performance comparison over various structural constraints.}
	\label{fig:hiloss}
\end{figure}

\vspace{2em}
\noindent\textbf{Study on Distributional Upsampling}

\noindent To examine the effectiveness of the distributional upsampling module, we compared SRResNet with UrbanFM-ne (using distributional upsampling but no external factors) and UrbanFM-sl (using structural loss instead of distributional upsampling), as shown in Figure~\ref{fig:hiloss}. In both $M$-$F$ settings, it can be seen that UrbanFM-sl regularized by $L_s$ performs very close to the SRResNet which is not constrained at all. Though under the setting of 16-64, Urban-sl achieves a smaller error than SRResNet in a subtle way, under the 16-128 setting they behave the opposite. On the contrary, UrbanFM-ne consistently outperforms the others on all three metrics. This result verifies that the distributional upsampling module is the better choice for imposing the structural constraint than using $L_s$.

\vspace{1mm}
\noindent\textbf{Study on External Factor Fusion}

\noindent External impacts, though are complicated, can assist the network for better inferences when they are properly modeled and integrated, especially in a more difficult situation when there is less data budget. Thereby, we study the effectiveness of external factors under four difficulties by randomly subsampling different ratio of the original training set, which are 10\%, 30\%, 50\% and 100\%, corresponding to four levels: hard, semi-hard, medium and easy.

\begin{figure}[h!]
	\centering
	\begin{subfigure}[b]{0.225\textwidth}
		\includegraphics[width=\textwidth]{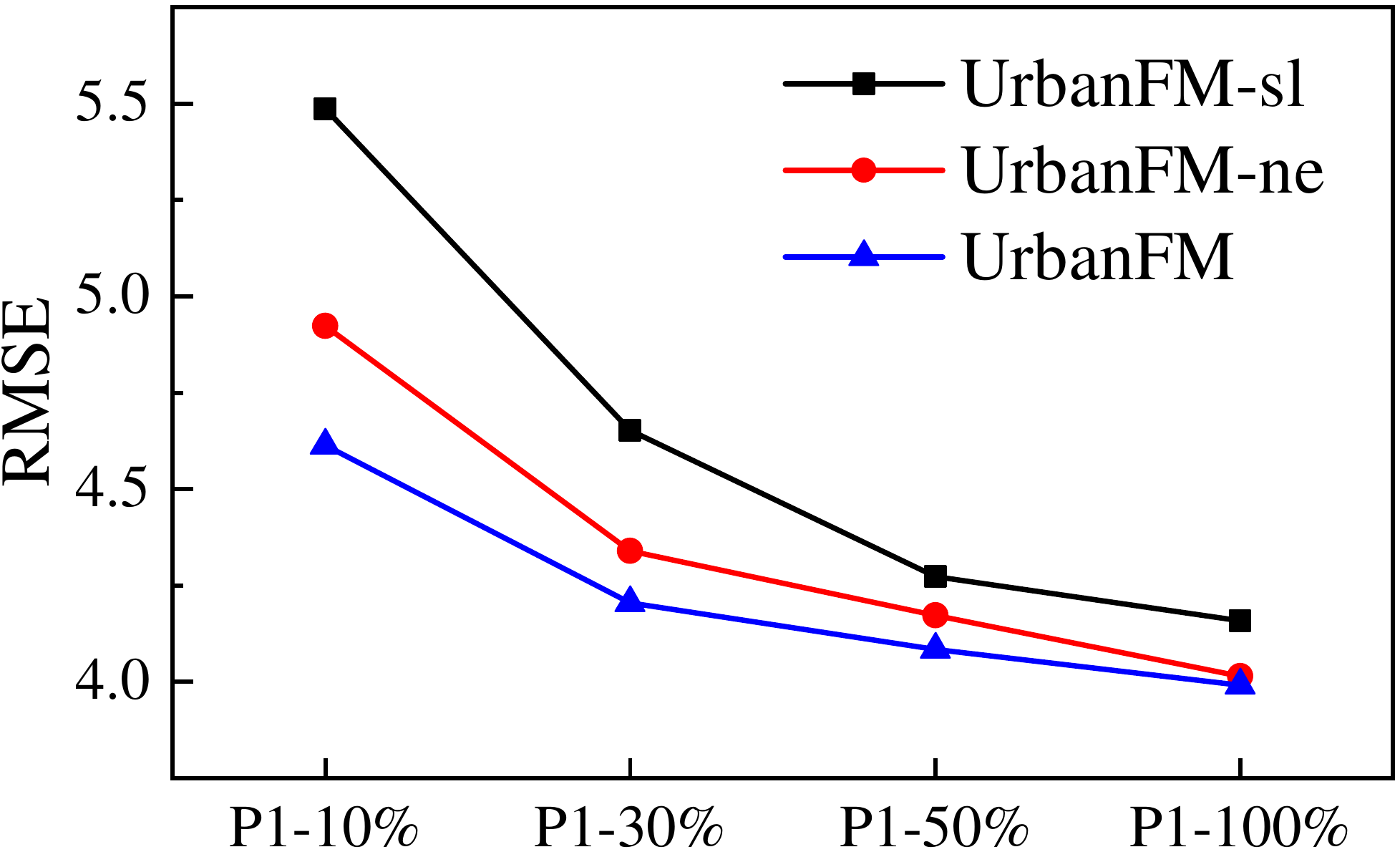}	
		\label{fig:ext_rmse}
	\end{subfigure}
	\hspace{1mm}
	\begin{subfigure}[b]{0.225\textwidth}
		\includegraphics[width=\textwidth]{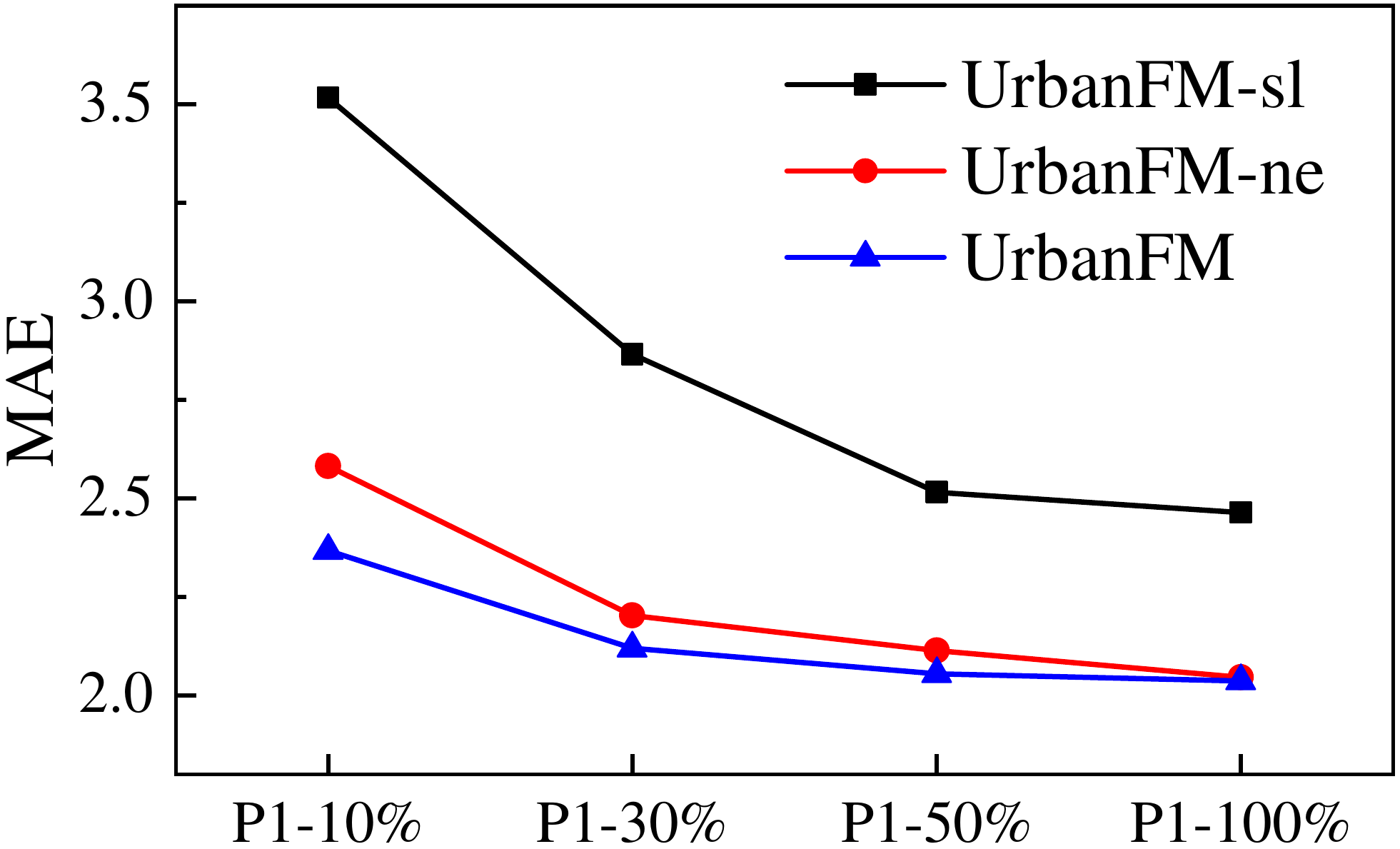}
		\label{fig:ext_mae}
	\end{subfigure}
	\vspace{-1em}
	\caption{Effects of external factors on four difficulties.}
	\label{fig:ext_effect}
\end{figure}

As shown in Figure~\ref{fig:ext_effect}, the \textit{gap} between UrbanFM and UrbanFM-ne becomes larger as we reduce the number of training data, indicating that external factor fusion plays a more important role in providing a priori knowledge. When the training size grows, the weight for the priori knowledge decreases, as there exists overlaying information between observed traffic flow and external factors. Thus, the network may learn to capture some external impacts by providing enough data. Moreover, this trend also occurs between UrbanFM and UrbanFM-sl, which illustrates that the $N^2$-\textit{Normalization} layer provides a strong structural prior to facilitate the network training.

%

\begin{table}[!b]
  \centering
  \caption{Results for different $M$-$F$ settings.}
  \vspace{-1em}
    \begin{tabular}{c|cc|rrr}
	\shline
    \multicolumn{1}{c|}{Methods} & \multicolumn{1}{l}{Settings} & \multicolumn{1}{l|}{\#Params} & \multicolumn{1}{l}{RMSE} & \multicolumn{1}{l}{MAE} & \multicolumn{1}{l}{MAPE} \\
	\hline
    SRResNet & 20-64 & 1.8M  & 4.317 & 2.586 & 0.725 \\
    UrbanFM & 20-64 & 1.9M  & 4.094 & 2.101 & 0.321 \\
    \hline
    SRResNet & 16-64 & 1.5M  & 4.261 & 2.520 & 0.689 \\
    UrbanFM & 16-64 & 1.7M  & 4.107 & 2.118 & 0.322 \\
    \hline
    SRResNet & 16-256 & 24.2M & 4.178 & 2.418 & 0.614 \\
    UrbanFM & 16-256 & 24.4M & 4.068 & 2.087 & 0.316 \\
    \shline
    \end{tabular}%
  \label{tab:params}%
\end{table}%

\vspace{1mm}
\noindent\textbf{Study on Parameter Size}

\noindent Table~\ref{tab:params} compares the average performance over P1 to P4. Across all hyperparameter settings, UrbanFM consistently outperforms SRResNet, advancing by at least 2.6\%, 13.7\% and 48.6\%. Besides, this experiment reveals that adding more ResBlocks (larger $M$) or increasing the number of filters (larger $F$) can improve the model performance. However, these also increase the training time and memory space. Considering the tradeoff between training cost and performance, we set the default setting of UrbanFM to be 16-128.

\vspace{1mm}
\noindent\textbf{Study on Efficiency}

\begin{figure}[!h]
	\centering
	\vspace{-1em}
	\begin{subfigure}[b]{0.225\textwidth}
		\includegraphics[width=\textwidth]{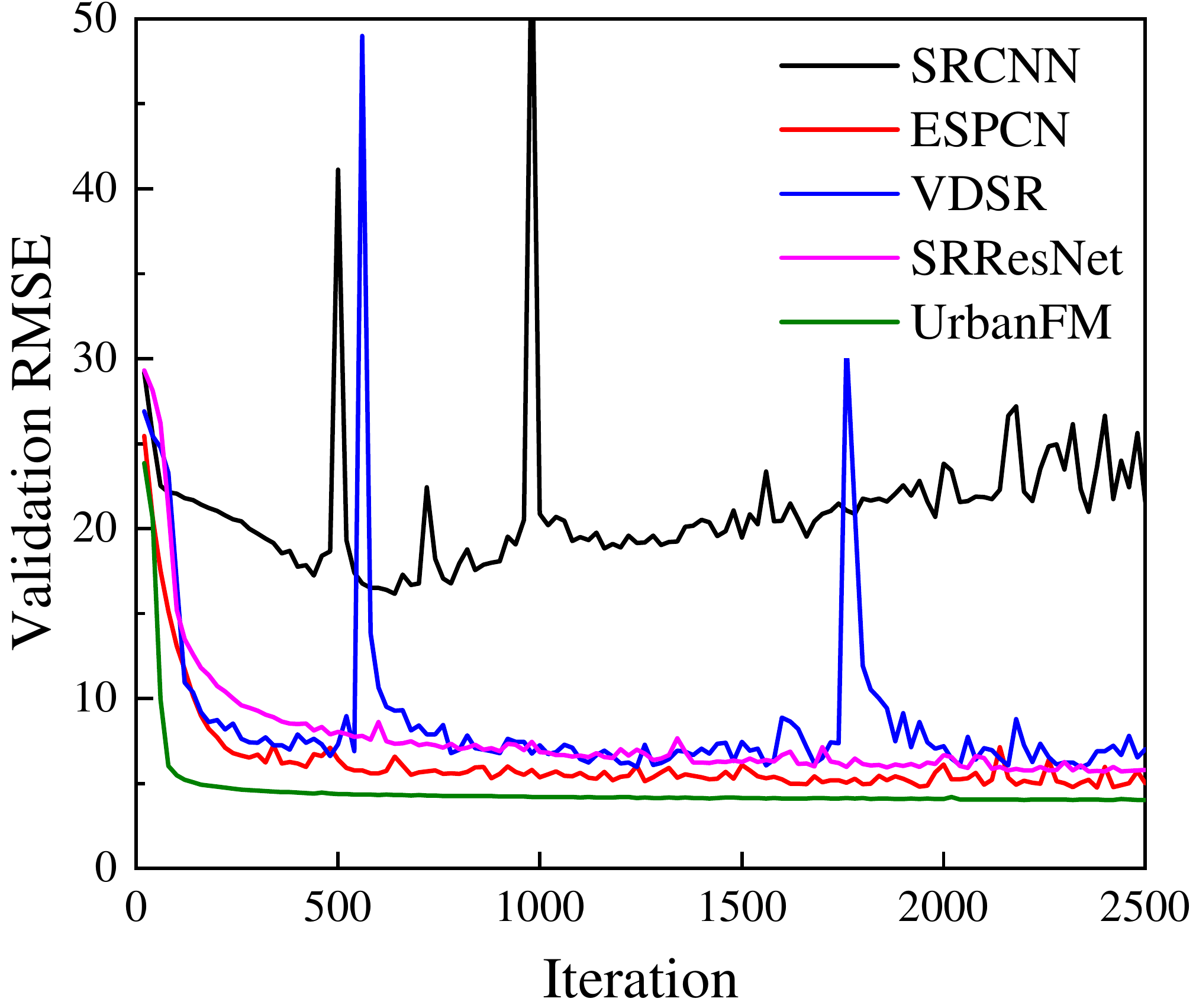}	
		\caption{\label{fig:effi1}Model comparison.}
	\end{subfigure}
	\hspace{1mm}
	\begin{subfigure}[b]{0.220\textwidth}
		\includegraphics[width=\textwidth]{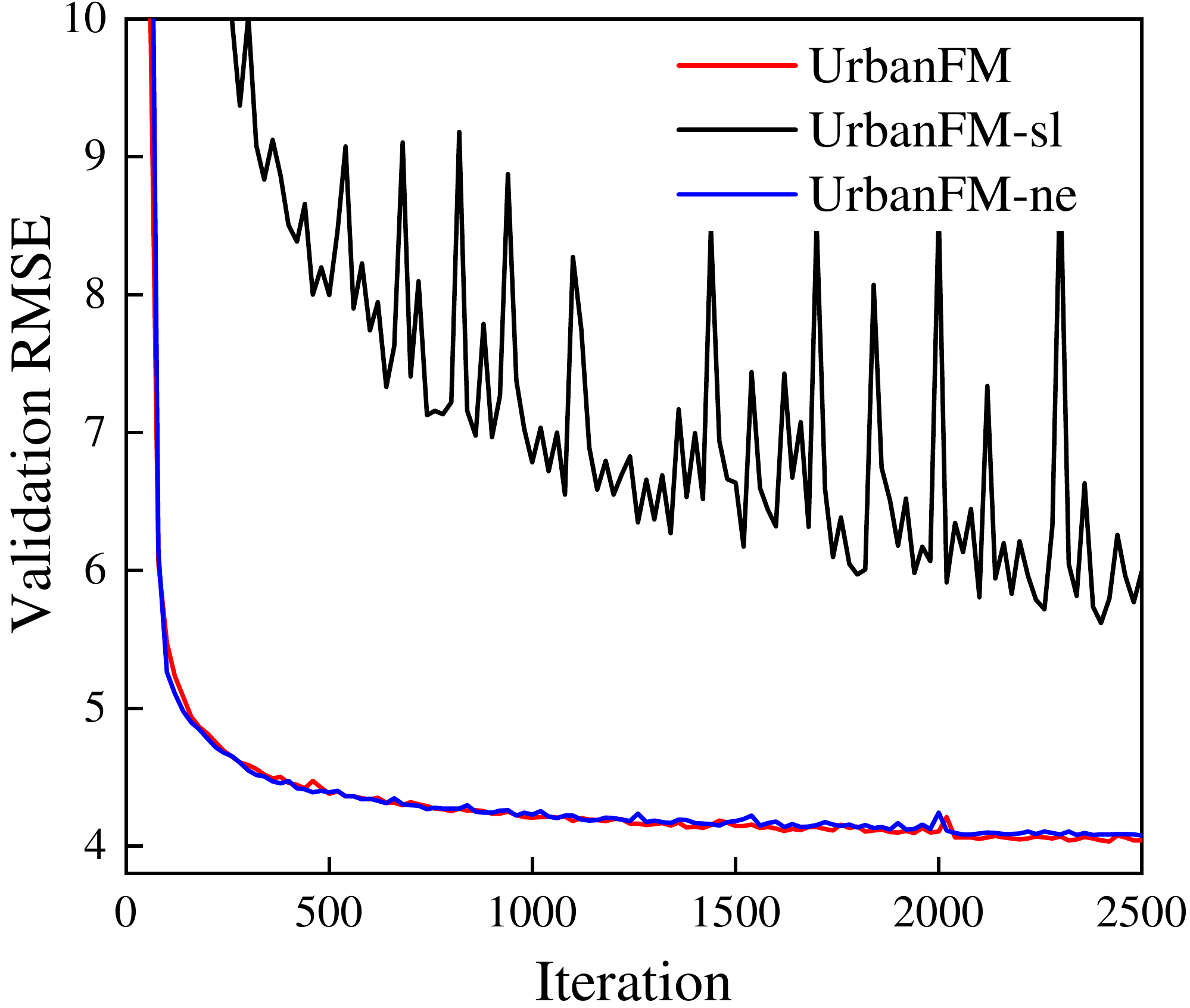}
		\caption{\label{fig:effi2}Variant comparison.}
	\end{subfigure}
	\vspace{-1em}
	\caption{Convergence speed of various methods.}
	\label{fig:efficiency}
\end{figure}

\noindent Figure~\ref{fig:efficiency} plots the RMSE on the validation set during the training phase using P1-100\%. Figure~\ref{fig:efficiency}(a) and \ref{fig:efficiency}(b) delineate that UrbanFM converges much \textit{smoother} and \textit{faster} than baselines and the variants. Specifically, \ref{fig:efficiency}(b) suggests such efficiency improvement can be mainly attributed to the $N^2$-\textit{Normalization} layer since UrbanFM-sl converges much slower and fluctuates drastically even it is constrained by $L_s$, when compared with UrbanFM and UrbanFM-ne. This also suggests that learning the spatial correlation is a non-trivial task. Moreover, UrbanFM-ne behaves closely to UrbanFM as external factors fusion affects the training speed subtly when training data are abundant as suggested by the previous experiments.

\vspace{1mm}
\noindent\textbf{Visualization}

\begin{figure}[!t]
	\centering
	\includegraphics[width=0.475\textwidth]{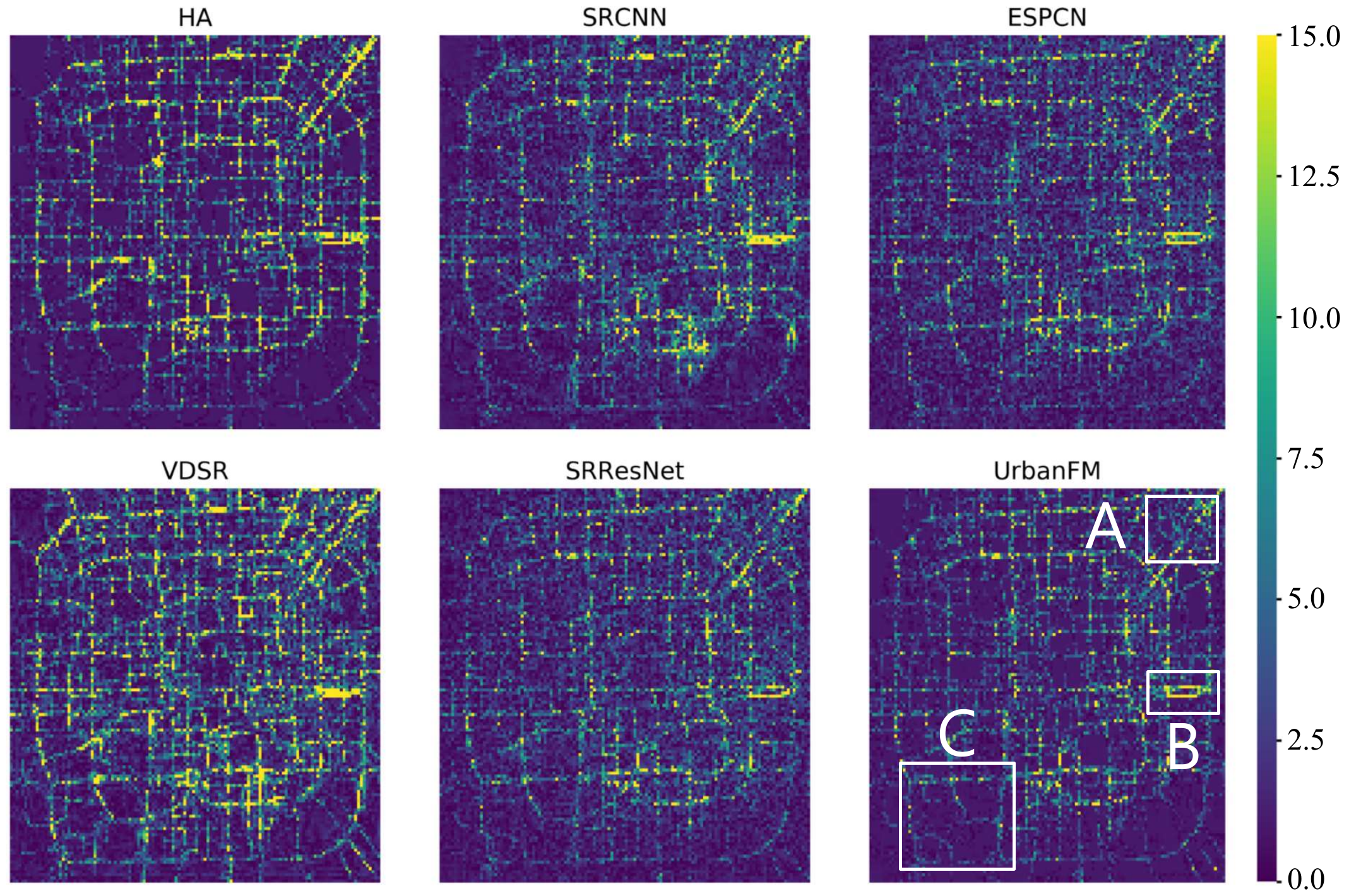}
	\vspace{-1em}
	\caption{\label{fig:diff} Visualization for inference errors among different methods. Best view in color.}	
	\vspace{-1em}
\end{figure}

\noindent1)~\textit{Inference error.} Figure~\ref{fig:diff} displays the inference error $\lVert \mathbf{X}^f-\tilde{\mathbf{X}^f} \rVert_{1,1}$ from UrbanFM and the other three baselines for a sample, where a brighter pixel indicates a larger error. Contrast with the other methods, UrbanFM achieves higher fidelity for totality and in details. For instance, area A and B are "hard area" to be inferred, as A (Sanyuan bridge, the main entrance to downtown) and B (Sihui bridge, a huge flyover) are two of the top congestion points in Beijing. Traffic flow of these locations usually fluctuates drastically and quickly, resulting in higher inference errors. Nonetheless, UrbanFM remains to produce better performance in these areas. Another observation is that the SR methods (SRCNN, ESPCN, VDSR and SRResNet) tend to generate blurry images as compared to structural methods (HA and UrbanFM). For instance, even if there is zero flow in area C, SR methods still generate error pixels as they overlap the predicted patches. This suggests the FUFI problem does differ from the ordinary SR problem and requires specific designs.

\begin{figure}[!t]
	\centering
	\includegraphics[width=0.45\textwidth]{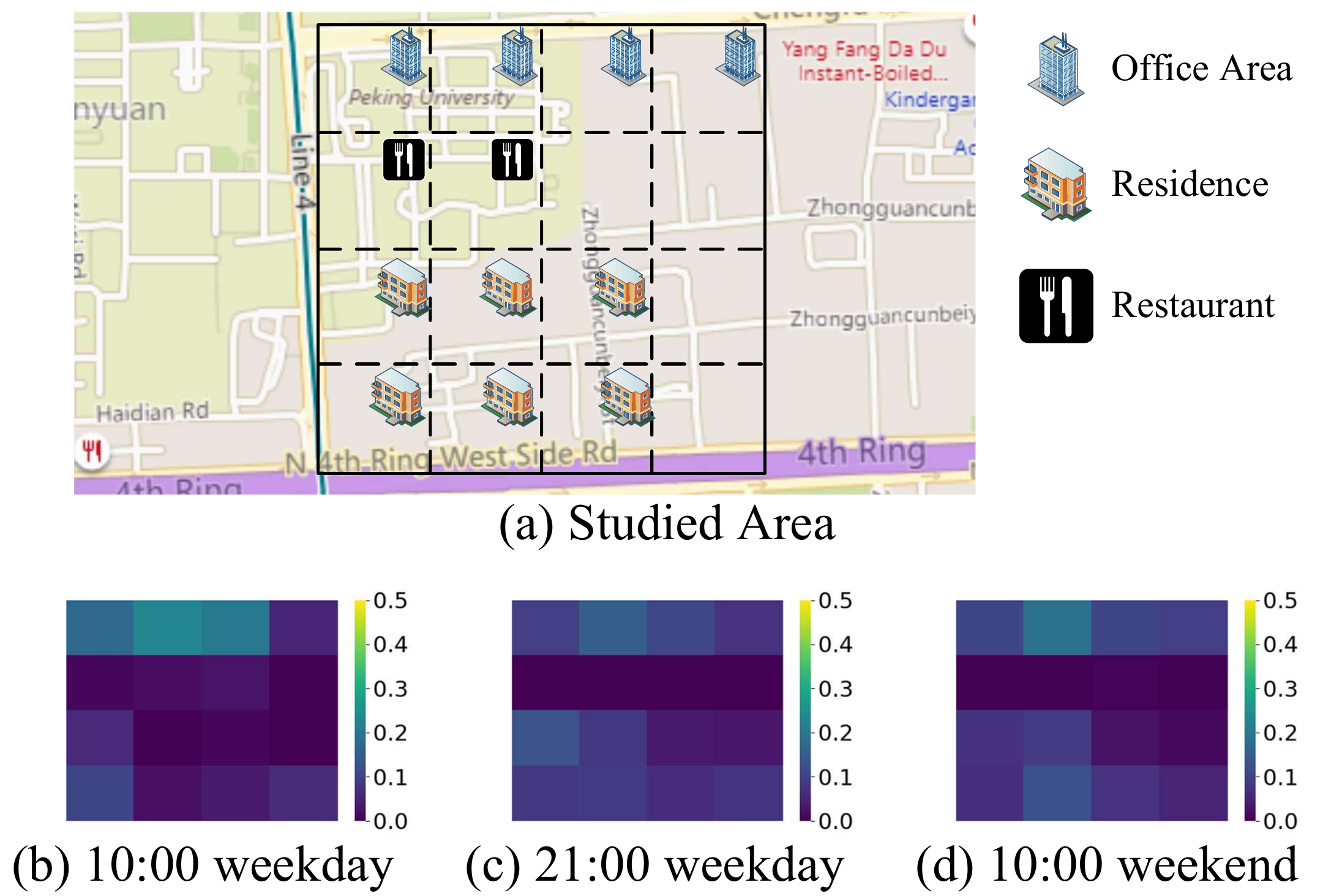}
	\vspace{-1em}
	\caption{\label{fig:case_study} Case study on a superregion near Peking Univ. See our github for further dynamic analysis on this area.}	
	\vspace{-1em}
\end{figure}

\noindent2)~\textit{External influence.} Figure~\ref{fig:case_study}(a)-(d) portray that the \textit{inferred} distribution over subregions varies along with external factor changes. On weekdays, at 10 a.m., people had already flowed to the office area to start their work (b); at 9 p.m., many people had returned home after a hard-working day (c). On weekends, most people stayed home at 10 a.m. but some industrial researchers remained working in the university labs. This result proves that UrbanFM indeed captures the external influence and learns to adjust the inference accordingly.

\subsection{Results on HappyValley}
Table~\ref{tab:hv} shows model performances using the HappyValley dataset. Note that in this experiment, we do not include DeepSD, since this task contains only $2\times$ upscaling and the DeepSD degrades to SRCNN in this case. One important trait of the HappyValley dataset is that it contains more spikes on the fine-grained flow distribution, which results in a much larger RMSE score versus that in the TaxiBJ task. Nonetheless, UrbanFM remains the winner method outperforming the best baseline by 3.5\%, 7.8\% and 22\%, and the UrbanFM-ne still holds the runner-up position. This proves that UrbanFM not only works on the large-scale scenario, but is also adaptable to smaller areas, which concludes our empirical studies.

\begin{table}[htbp]
  \centering
  \caption{Results comparison on Happy Valley.}
  \vspace{-0.8em}
    \begin{tabular}{c|cc|ccc}
    \shline
    Methods & Settings & \#Params & RMSE  & MAE   & MAPE \\
    \hline
    MEAN  & x     & x     & 9.206 & 2.269 & 0.799 \\
    HA    & x     & x     & 8.379 & 1.811 & 0.549 \\
    \hline
    SRCNN & 768   & 7.4M & 8.291 & 2.175 & 0.816 \\
    ESPCN & 768   & 7.5M & 8.156 & 2.155 & 0.805 \\
    VDSR  & 20-64 & 0.6M & 8.490 & 2.128 & 0.756 \\
    SRResNet & 16-128 & 5.5M & 8.318 & 1.941 & 0.679 \\
    \hline
    UrbanFM-sl & 16-128 & 5.5M & 8.312 & 1.939 & 0.677 \\
    UrbanFM-ne & 16-128 & 5.5M & 8.138 & 1.816 & 0.537 \\
    \textbf{UrbanFM} & \textbf{16-128} & \textbf{5.6M} & \textbf{8.030} & \textbf{1.790} & \textbf{0.531} \\
    \shline
    \end{tabular}%
  \label{tab:hv}%
\end{table}%

\section{Related Work}

\subsection{Image Super-Resolution}\label{sec:SISR}
Single image super-resolution (SISR), which aims to recover a high-resolution (HR) image from a single low-resolution (LR) image, has gained increasing research attention for decades. This task finds direct applications in many areas such as face recognition \cite{gunturk2003eigenface}, fine-grained crowdsourcing \cite{thornton2006sub} and HDTV \cite{park2003super}. Over years, many SISR algorithms have been developed in the computer vision community. To tackle the SR problem, early techniques focused on interpolation methods such as bicubic interpolation and Lanczos resampling \cite{duchon1979lanczos}. Also, several studies utilized statistical image priors \cite{sun2008image, tai2010super} to achieve better performances. Advanced works aimed at learning the non-linear mapping between LR and HR images with neighbor embedding \cite{chang2004super} and sparse coding \cite{yang2010image,timofte2014a}. However, these approaches are still inadequate to reconstruct realistic and fine-grained textures of images.

Recently, a series of models based on deep learning have achieved great success in terms of SISR since they do not require any human-engineered features and show the state-of-the-art performance. Since \citeauthor{dong2016srcnn} \cite{dong2016srcnn} first proposed an end-to-end mapping method represented as CNNs between the low-resolution (LR) and high-resolution (HR) images, various CNN based architectures have been studied for SR. Among them, \citeauthor{shi2016espcn} \cite{shi2016espcn} introduced an efficient sub-pixel convolutional layer which is capable of recovering HR images with very little additional computational cost compared with the deconvolutional layer at training phase. Inspired by VGG-net for ImageNet classification \cite{simonyan2014vgg}, a very deep CNN was applied for SISR in \cite{kim2016vdsr}. However, training a very deep network for SR is really hard due to the small convergence rate. \citeauthor{kim2016vdsr} \cite{kim2016vdsr} showed residual learning speed up their training phase and verified that increasing the network depth could contribute to a significant improvement in SR accuracy.

Despite good performance on the RMSE accuracy, the generated image remains smooth and blurry. To address this problem, \citeauthor{ledig2017srgan} \cite{ledig2017srgan} first proposed a perceptual loss function which consists of an adversarial loss to push their solution to the natural image manifold, and a content loss for the better reconstruction of high-frequency details. \citeauthor{lim2017enhanced} \cite{lim2017enhanced} developed an enhanced deep SR network that shows the state-of-the-art performance by removing unnecessary modules in \cite{ledig2017srgan}. Apart from super-resolving classical images, there are limited studies that focus on utilizing super-resolution methods to solve real-world problems in the urban area. For example, \citeauthor{vandal2017deepsd} \cite{vandal2017deepsd}, presented a stacked SRCNN \cite{dong2016srcnn} for statistical downscaling of climate and earth system simulations based on observational and topographical data.

However, these approaches are not suitable for the FUFI problem since the flow data present a very specific structural constraint with regard to natural images, as such, the related arts cannot be simply applied to our application in terms of efficiency and effectiveness. To the best of our knowledge, \emph{we are the first to formulate and solve the problem for fine-grained urban flow inference}.

\subsection{Urban Flows Analysis}
Due to the wide applications of traffic analysis and the increasing demand for real-time public safety monitoring, urban flow analysis has recently attracted the attention of a large amount of researchers \cite{zheng2014urban}. Over the past years, \citeauthor{zheng2014urban} \cite{zheng2014urban} first transformed public traffic trajectories into other data formats, such as graphs and tensors, to which more data mining and machine learning techniques can be applied. Based on our observation, there were several previous works \cite{song2014prediction,fan2015citymomentum} forecasting millions, or even billions of individual mobility traces rather than aggregated flows in a region.

Recently, researchers have started to focus on city-scale traffic flow prediction \cite{hoang2016fccf}. Inspired by deep learning techniques that power many applications in modern society \cite{lecun2015deep}, a novel deep neural network was developed by \citeauthor{zhang2016dnn} \cite{zhang2016dnn} to simultaneously model spatial dependencies (both near and distant), and temporal dynamics of various scales (\emph{i.e.}, closeness, period and trend) for citywide crowd flow prediction. Following this work, \citeauthor{zhang2017deep} \cite{zhang2017deep} further proposed a deep spatio-temporal residual network to collectively predict inflow and outflow of crowds in every city grid. To address the data scarcity issue in crowd flows, very recent study \cite{wang2018crowd} aims to transfer knowledge between different cities to help target city learn a better prediction model from the source city. Apart from the above applications, we aim to solve a novel problem (FUFI) on urban flows in this study.
\section{Conclusion}
In this paper, we have formalized the fine-grained urban flow inference problem and presented a deep neural network-based method (UrbanFM) to solve it. UrbanFM has addressed the two challenges that are specific to this problem, \emph{i.e.}, the spatial correlation as well as the complexities of external factors, by leveraging the original distributional upsampling module and the external factor fusion subnet. Experiments have shown that our approach advances baselines by at least 4.5\%, 17.0\% and 54.1\% on TaxiBJ dataset and 3.5\%, 7.8\% and 22\% on HappyValley dataset in terms of the three metrics. Various empirical studies and visualizations have confirmed the advantages of UrbanFM on both efficiency and effectiveness.

In the future, we will explore more on improving the model structure, and pay more attention to reducing errors in hard regions.
\appendix

\bibliographystyle{ACM-Reference-Format}
\bibliography{references}

\newpage
\appendix
\section{Appendix for Reproducibility}
To support the reproducibility of the results in this study, we have released our code and data\footnote{https://github.com/yoshall/UrbanFM}. Our implement is based on Pytorch 0.4.1. Here, we present the details of the dataset, normalization method and experimental settings.

\subsection{Statistics of Datasets}
In section \ref{sec:exp}, we have illustrated how we split the training, validation and test set based on the two real-world datasets: TaxiBJ and HappyValley. Since there are some coarse-grained data with most zero entries (\emph{i.e.}, extremely noisy data), we directly remove them from the original dataset. Here, we display the details of available samples in each set in Figure \ref{tab:partition}.

\begin{table}[htbp]
\centering
\tabcolsep=0.28cm
\caption{The details of partition over two datasets}
\vspace{-1em}
    \begin{tabular}{c|c|ccc}
    \shline
    \multirow{2}*{Dataset} & \multirow{2}*{Time Span} & \multicolumn{3}{c}{Size} \\
\cline{3-5}          &       & train & valid & test \\
    \hline
    \multirow{4}[1]{*}{TaxiBJ} & P1    & 1530  & 765   & 765 \\
        & P2    & 1779  & 889   & 891 \\
        & P3    & 1746  & 873   & 873 \\
        & P4    & 2122  & 1061  & 1061 \\
    \hline
    HappyValley & x     & 2188  & 273   & 275 \\
    \shline
    \end{tabular}%
\label{tab:partition}%
\end{table}%

\subsection{Normalization Method}
\begin{figure}[!h]
	\centering
	\includegraphics[width=0.47\textwidth]{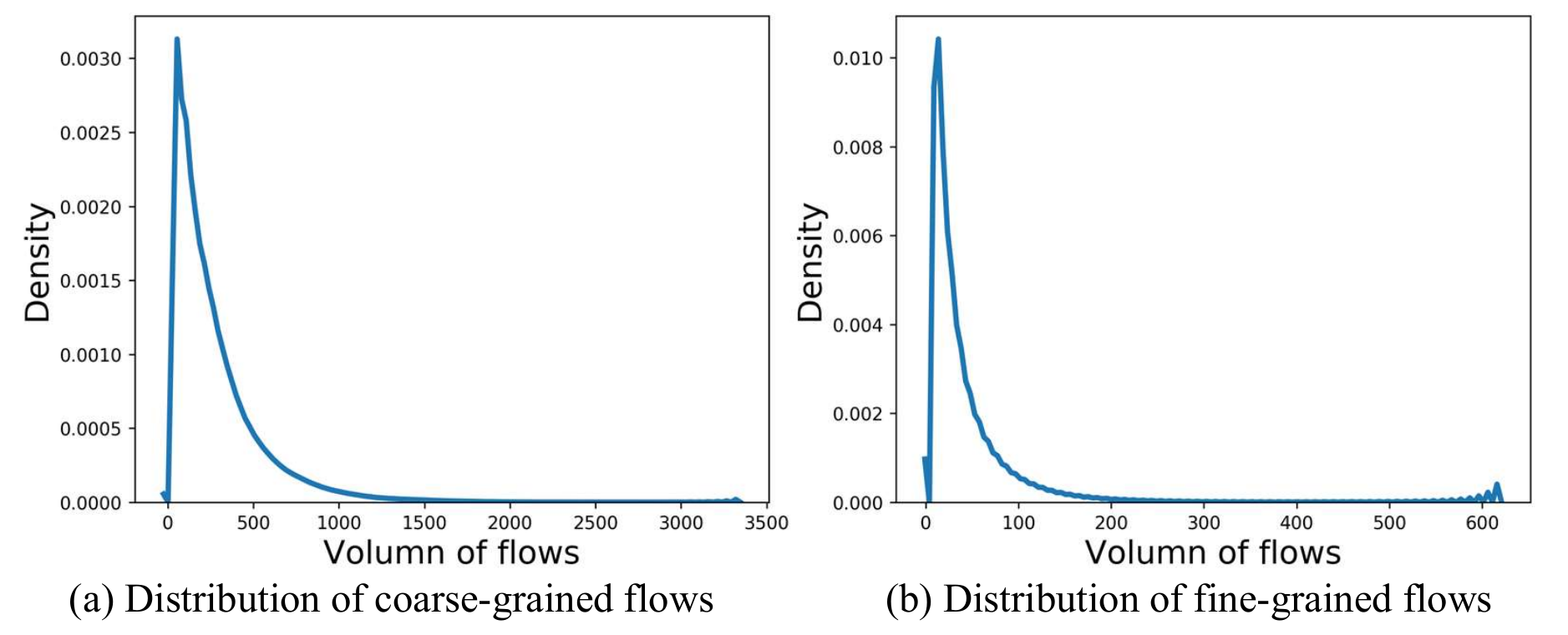}
    \caption{Distribution of urban flows in TaxiBJ dataset.}
    \vspace{-1em}
    \label{fig:norm}
\end{figure}

We employ data normalization before the training phase to speed up the convergence of our method. Recall that we obtain the inferred distribution $\mathbf{H}^f_\pi$ (in the range $[0, 1]$) together with $\mathbf{X}^c_{up}\in\mathbb{R}^{NI\times NJ}_{+}$ which is upsampled from the original coarse-grained flow map. Every entry of the final fine-grained output $\tilde{\mathbf{X}}^f=\mathbf{X}^c_{up}\odot\mathbf{H^f_\pi}$ is positive, \emph{i.e.}, $\tilde{x}^f_{i',j'}>0$. Hence, we use Min-Max normalization to scale the input and the output to $[0,1]$. Since the original scale of coarse- and fine-grained flows are different, we plot each regional flow volumes from the flow maps of TaxiBJ in Figure \ref{fig:norm}, where a long tail can be observed in both settings. An explanation is that some regions can sometimes witness a high flow volume, which can be attributed to the rush hours or traffic jams\cite{zheng2014urban}. Due to the long tail, suppose we simply use the maximum of such flows as max-scaler, it will restrict most values to be much smaller than 1. Based on this observation, we set the two max-scaler 1500 and 100 in coarse- and fine-grained data respectively. Likewise, we use the same method to decide the proper scaler in HappyValley dataset.

\subsection{Detailed Settings of Baselines}
We detail the model configuration as well as hyperparameter searching spaces for each baseline in this section.
\begin{itemize}[leftmargin=*]
  \item \textbf{Mean Partition (Mean)}: It is parameter-free and can be directly applied on the test set.
  \item \textbf{Historical Average (HA)}: Firstly, we compute the mean distribution matrix on the training set (with no parameters). Then, the matrix is applied to generate the fine-grained flow map over a coarse-grained observation. 
  \item \textbf{SRCNN}: Since SRCNN under its default setting achieves inferior performance and takes a long time to converge, we test different hyperparameters for it, so as to find the best setting. Suppose there are $F_1$ and $F_2$ filters in the two convolutional layers of such method. We conduct a grid search over $F_1=\{64, 128, 256, 512, 768, 1024\}$ and $F_2=\{32, 64, 128, 256, 384, 512\}$. The setting in which $F_1=768$ and $F_2=384$ outperforms the others in the validation set.
  \item \textbf{ESPCN}: Similar to SRCNN with three-staged architecture, we leverage $F_1=768$ and $F_2=384$ as the number of filters in different convolutional layers respectively.
  \item \textbf{DeepSD}: Experiments show that ESPCN is more efficient and effective than SRCNN \cite{shi2016espcn}. In order to speed up this method based on stacked SRCNNs, we use ESPCN to replace the SRCNN in the original paper, (\emph{i.e.}, a stacked ESPCN).
  \item \textbf{VDSR}: The depth of convolutional blocks $D$ and the number of filters $F$ in convolutional layer are two main hyperparameters. We utilize the default setting $D=20$ and $F=64$ as suggested by the authors \cite{kim2016vdsr}.
  \item \textbf{SRResNet} \cite{ledig2017srgan}: In our paper, we compare our method with SRResNet from multiple angles. There are two main hyperparameters in SRResNet, including the depth of residual blocks $M$ and number of filters in convolutional layer $F$. We test $M = \{ 16, 20\}$ and $F=\{64, 128, 256\}$ in different experiments, which is detailed in Section \ref{sec:exp}.
\end{itemize}

\subsection{Detailed Settings of UrbanFM}
We first introduce how we implement $N^2$-Normalization layer based on Pytorch, and further present the detailed settings of two main components of our approach, \emph{i.e.}, inference network and external factor fusion subnet.

\subsubsection{$N^2$-Normalization Layer}
\begin{figure}[!h]
	\centering
	\includegraphics[width=0.47\textwidth]{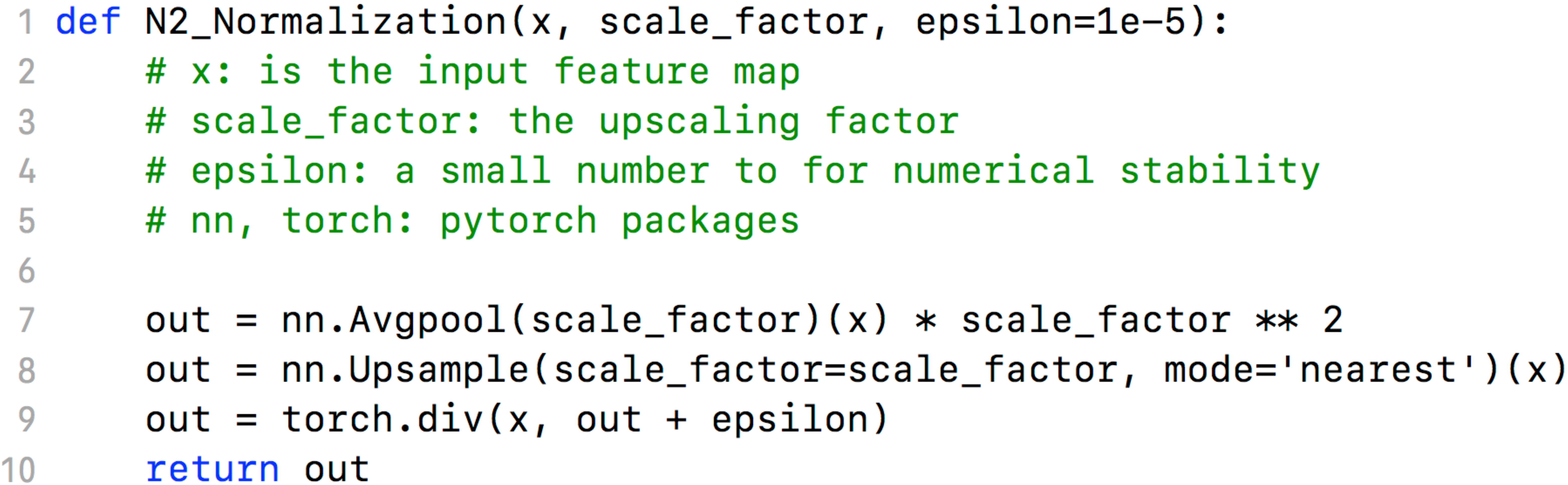}
	\caption{\label{fig:n2norm} Implementation of $N^2$-Normalization layer based on PyTorch 0.4.1}	
\end{figure}

Figure~\ref{fig:n2norm} illustrates the Pytorch implementation of  $N^2$-Normalization layer, which plays a significant role in our method.

\subsubsection{Inference Network} Table \ref{tab:infnet} show the detailed configuration for the inference network which is depicted in Figure \ref{fig:framework} (from left to right). Note that the upscaling factor $N=4$ in this example.
\begin{table}[!ht]
    \centering
    \caption{Details settings of Inference Network in Figure \ref{fig:framework}, where settings k-s-n means the size of kernel, stride and number of filters in a certain convolutional layer. We omit the batch size in the format of output for simplicity.}
      \begin{tabular}{l|c|c}
      \shline
      Layer  & Setting & Output \\
      \hline
      Concat\_1 & x     & $2 \times I \times J$ \\
      Conv\_1 & 9-1-64 & $64 \times I \times J$ \\
      ReLU  & x     & $64 \times I \times J$ \\
      \hline
      Conv\_1 (ResBlock) & 3-1-$F$ & $F \times I \times J$ \\
      BatchNorm\_1 (ResBlock) & x     & $F \times I \times J$ \\
      ReLU (ResBlock) & x     & $F \times I \times J$ \\
      Conv\_2 (ResBlock) & 3-1-$F$ & $F \times I \times J$ \\
      BatchNorm\_2 (ResBlock) & x     & $F \times I \times J$ \\
      Other ResBlocks &  ...     & ... \\
      \hline
      Conv\_2 & 3-1-$F$ & $F \times I \times J$ \\
      BatchNorm & x     & $F \times I \times J$ \\
      \hline
      Conv (SubPixel Block\_1) & 3-1-4$F$ & $4F \times I \times J$ \\
      BatchNorm (SubPixel Block\_1) & x     & $4F \times I \times J$ \\
      PixelShuffle (SubPixel Block\_1) & x     & $F \times 2I \times 2J$ \\
      ReLU (SubPixel Block\_1) & x     & $F \times 2I \times 2J$ \\
      Conv (SubPixel Block\_2) & 3-1-4$F$ & $4F \times 2I \times 2J$ \\
      BatchNorm (SubPixel Block\_2) & x     & $4F \times 2I \times 2J$ \\
      PixelShuffle (SubPixel Block\_2) & x     & $F \times 4I \times 4J$ \\
      ReLU (SubPixel Block\_2) & x     & $F \times 4I \times 4J$ \\
      Concat\_2 & x     & $(F+1) \times 4I \times 4J$ \\
      Conv\_3 & 9-1-1 & $1 \times 4I \times 4J$ \\
      $N^2$-Normalization & x     & $1 \times 4I \times 4J$ \\
      \shline
      \end{tabular}%
    \label{tab:infnet}%
  \end{table}%
  
\subsubsection{External Factor Fusion Subnet}
Before inputting to the subnet, we the use embedding method to convert the categorical features (like day of week, weather condition\footnote{TaxiBJ witnesses 16 kinds of weather conditions: Sunny, Cloudy, Overcast, Rainy, Sprinkle,  ModerateRain,  HeavyRain, Rainstorm, Thunderstorm, FreezingRain, Snowy,  LightSnow, ModerateSnow, HeavySnow, Foggy, Sandstorm, Dusty. For HappyValley, only 8 types of above weather conditions are included.}) to learned representations respectively, \emph{i.e.}, real-valued vectors. As shown in Table~\ref{tab:embed}, we detail the embedding settings for each external factor.

\begin{table}[!htbp]
    \centering
    \caption{Embedding setting of external factors.}
      \begin{tabular}{l|l|c|c}
      \shline
      Data  & Feature & \#Categroies & Embed Length\\
      \hline
      \multirow{3}*{Meteorology} & Temperature & x     & 1 \\
  \cline{2-4}          & Wind speed & x     & 1 \\
  \cline{2-4}          & Weather & 16 (8) & 3 \\
      \hline
      \multirow{4}{*}{Time} & Holiday & 2     & 1 \\
  \cline{2-4}          & Weekend & 2     & 1 \\
  \cline{2-4}          & Day of week & 7     & 2 \\
  \cline{2-4}          & Hour of day & 24    & 3 \\
      \hline
      Event & Ticket price & x     & 1 \\
      \shline
      \end{tabular}%
    \label{tab:embed}%
  \end{table}%

  Table \ref{tab:extfusion} shows the details of the external subnet. The settings of dense layer denotes the number of hidden units, while that of dropout layer represents its randomly dropping rate. It is also illustrated from left to right of Figure \ref{fig:framework}.

  \begin{table}[!ht]
      \centering
      \tabcolsep=0.3cm
      \caption{Details of External Factor Fusion in Figure \ref{fig:framework}.}
        \begin{tabular}{l|c|c}
        \shline
        Layer  & Setting & Output \\
        \hline
        Dense\_1 & 128    & $128$ \\
        Dropout & 0.3 & $128$ \\
        ReLU\_1  & x     & $128$ \\
        Dense\_2 & $I \times J$    & $I \times J$ \\
        ReLU\_2  & x     & $I \times J$ \\
        \hline
        Conv (SubPixel Block\_1) & 3-1-4 & $4 \times I \times J$ \\
        BatchNorm (SubPixel Block\_1) & x     & $4 \times I \times J$ \\
        PixelShuffle (SubPixel Block\_1) & x     & $1 \times 2I \times 2J$ \\
        ReLU (SubPixel Block\_1) & x     & $1 \times 2I \times 2J$ \\
        Conv (SubPixel Block\_2) & 3-1-4 & $4 \times 2I \times 2J$ \\
        BatchNorm (SubPixel Block\_2) & x     & $4 \times 2I \times 2J$ \\
        PixelShuffle (SubPixel Block\_2) & x     & $1 \times 4I \times 4J$ \\
        ReLU (SubPixel Block\_2) & x     & $1 \times 4I \times 4J$ \\
        \shline
        \end{tabular}%
      \label{tab:extfusion}%
    \end{table}%

\clearpage
\end{document}